\documentclass{article} % For LaTeX2e
\usepackage{iclr2027_conference,times}

\usepackage{amsmath,amsfonts,bm}
\usepackage{wrapfig}

\def\eqref#1{equation~\ref{#1}}
\def\1{\bm{1}}

\DeclareMathAlphabet{\mathsfit}{\encodingdefault}{\sfdefault}{m}{sl}
\SetMathAlphabet{\mathsfit}{bold}{\encodingdefault}{\sfdefault}{bx}{n}

\usepackage{hyperref}
\usepackage{url}
\usepackage{booktabs} 
\usepackage{amssymb}
\usepackage{amsmath}
\usepackage{amsfonts}
\usepackage[version=4]{mhchem}
\usepackage{stmaryrd}
\usepackage{caption}
\usepackage{graphicx}
\usepackage[table]{xcolor}
\usepackage{xspace}

\title{Attention Sinks in VLMs Are Borrowed, \\ Not Seen}

\author{Dengjia Zhang$^{1,2}$ \quad Alexander Martin$^{1}$ \quad Pengyu Yan$^{4}$ \quad Xihan Sun$^{3}$  \quad Jiaqi Li$^{1}$ \\
\textbf{Benjamin Van Durme$^{1}$ \quad Andrew Yates$^{1,5}$ \quad Reno Kriz$^{1,5}$ \quad William Walden$^{1,5}$} \\[0.5em]
$^{1}$Johns Hopkins University
$^{2}$University of Pittsburgh \\
$^{3}$Vanderbilt University \quad
$^{4}$University at Buffalo \\
$^{5}$Human Language Technology Center of Excellence, Johns Hopkins University \\[0.5em]
\texttt{dzhang@jhu.edu}
}

\iclrfinalcopy % Uncomment for camera-ready version, but NOT for submission.

\newcommand{\vsink}{visual attention sink\xspace}

\newcommand{\tsink}{text-sink\xspace}

\newcommand{\vsinks}{visual attention sinks\xspace}
\newcommand{\Vsinks}{Visual attention sinks\xspace}
\newcommand{\tsinks}{text-sinks\xspace}

\DeclareUnicodeCharacter{00D7}{\ifmmode\times\else{$\times$}\fi}
\DeclareUnicodeCharacter{2192}{\ifmmode\rightarrow\else{$\rightarrow$}\fi}

\begin{document}

\maketitle

\begin{abstract}
% We explore how to find and mitigate visual attention sinks in vision-language models (VLMs). Visual attention sinks receive a large share of attention over the image regardless of the query or task, and they mislead methods that use attention maps as a record of where the model ``looked''. Prior work has detected sink through massive activation dimensions and proposed redistributing their attention, but no work has addressed what makes a token become a sink. We close the gap by examining the keys of visual attention sinks. We show that the keys of visual attention sinks carry a dominant component along the keys of \tsinks, reflected in a high centered cosine similarity between themselves and \tsinks while the normal visual attention sinks will not. Therefore, we introduce a training free method to clean visual attention sinks based on locating the image keys that have high cosine similarity to \tsink keys and resetting them their component to the median of ordinary image keys. The method adds negligible computation in a forward pass. As a demonstration, an entropy-triggered re-look that crops the most-attended region of the cleaned map improves accuracy on fine-detail benchmarks across model families without any training. Due to its negligible computational overhead, the cleaning step can further be incorporated into training pipelines such as reinforcement learning, where attention maps free of sinks may serve as a grounding signal without additional cost.

We study how to detect and mitigate visual attention sinks in vision-language models (VLMs). Visual attention sinks are image tokens that absorb a large share of attention over the image regardless of the query or task, and they mislead methods that treat attention maps as a record of where the model ``looked.'' Prior work has detected sinks through massive activation dimensions and proposed redistributing their attention, but none have addressed what makes a token become a sink. Comparing six candidate detectors on ten VLMs from four families, we find that only the centered cosine similarity between an image token's key and the key of the language model's \tsink detects visual sinks. Thus, a visual sink is an image token whose key aligns with the \tsink's key. Moving a sink's key component along this direction to the median of the image keys sends its attention back to the \tsink (\textbf{Are Borrowed}) while leaving model answers unchanged~(\textbf{Not Seen}). Building on this, we propose a training-free, entropy-triggered re-look that first cleans the attention map, then crops the most-attended region of the cleaned map and inserts it into the reasoning trace. Cleaning adds negligible computation to the forward pass, and the re-look improves accuracy with a mean gain of $+4.3$ points on fine-detail VQA benchmarks across model families.

\end{abstract}

\section{Introduction}
\label{sec:intro}

The attention map of a vision--language model is widely read as a record of where the model looks. It is used to trace how object information flows from image tokens to text \citep{kaduri2025s,neo2025towards,jiang2025devils} and to guide model behavior: cropping or highlighting the most-attended region \citep{zhang2025vicrop,shen2025zoomeye,liu2026seeing}, reading out grounding heads \citep{kang2025localization}, pruning the least-attended tokens \citep{chen2024fastv,zhang2025sparsevlm,yang2025visionzip}, and reweighting attention against hallucination \citep{huang2024opera,liu2024pai,che2025easy}. However, this record is often misleading as a few image tokens, the \vsinks \citep{kang2025see}, take a large share of the image attention regardless of the question. Any method that interprets or uses VLM attention can be misled by these tokens.

Attention sinks were first observed in language models as a few tokens that absorb much of the attention regardless of their content \citep{xiao2024streamingllm,su2026attention}. These tokens carry massive activations \citep{sun2024massive} and their keys act as a bias rather than as content \citep{gu2025sink,barbero2025first}. Every VLM inherits such \tsinks from its language model. \Vsinks have been identified through massive hidden dimensions \citep{kang2025see}, attributed to the vision encoder, language model, rotary position encoding \citep{darcet2024registers,luo2026sink,choi2026sinks,zhang2026drives}, and either removed or exploited by downstream methods \citep{li2026sinkpruner,shukla2026sage}. However, none of these studies has examined what a \vsink is or what makes an image token become one. Without this, it is unclear whether the sinks reported in different VLMs are the same phenomenon and whether they carry useful information.

We study \vsinks in ten VLMs from four families and find that they are identified by their keys. The features prior work associates with sinks, the hidden-state norm \citep{kang2025see} and the value and key norms \citep{sun2024massive}, each detect sinks in only some model families. Since a \tsink's key acts as a bias \citep{gu2025sink}, we build three detectors that test whether image sinks share this bias: the spread of the attention logits a token receives across queries, and the raw and centered cosine similarity between an image token's key and the text-sink key. Only the centered cosine detects sinks in every sink-bearing model (AUC $\geq 0.99$). Thus, a \vsink is an image token whose key points along the text-sink key.

\begin{figure}
    \centering
    \includegraphics[width=0.8\linewidth]{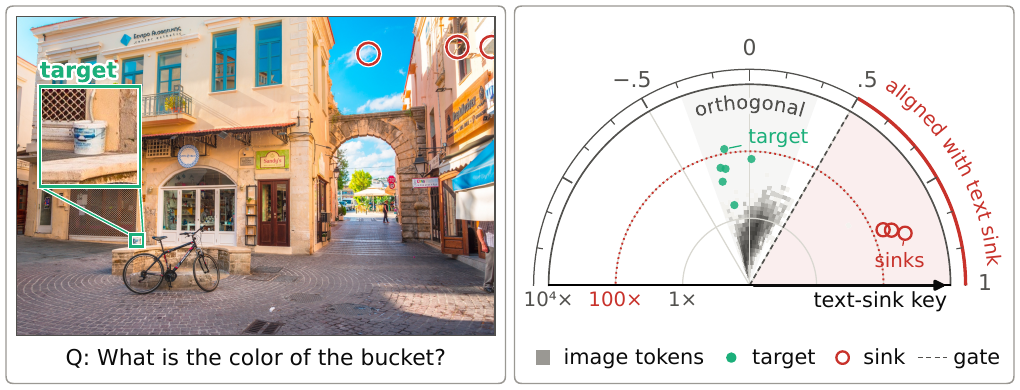}
  \caption{\textbf{Visual attention sinks point the same way as the \tsink.}
  Left: \vsinks (red, $\geq 100\times$ uniform attention outside the target) and the target (green) in a V$^\ast$ image.
  Right: key direction (angle) vs. attention (radius) of every image token}
    \label{fig:placeholder}
    \vspace{-1em}
\end{figure}

This definition also gives a way to remove \vsinks. At every layer, we flag the image tokens whose centered cosine with the text-sink key exceeds a threshold, and move the component of each flagged key along  the text-sink direction to the median over image keys, leaving queries and values unchanged. In most sink-bearing models, this removal reveals two properties. First, \vsinks \emph{are borrowed}. The attention the \vsinks held returns to the \tsink rather than spreading over the image. Second, \vsinks are \emph{not seen}. The model's output distribution barely changes, with low KL divergence from the original distribution. The attention map, however, does change, with the most-attended image token often landing on the queried object. 

With the \vsinks removed and the model's behavior preserved, the attention map is again a record of where the model looks. Reasoning VLMs increasingly revisit the image mid-trace by cropping or zooming, but they learn when and where to do so through reinforcement learning \cite{zheng2025deepeyes,su2025pixelreasoner,zhang2025cof,lai2025minio3} or rely on a separate search or tool calling procedure \citep{wu2024vstar, shen2025zoomeye,hu2024sketchpad}. We introduce a training-free re-look for revisiting the image only when the model becomes uncertain. At the first decoding step where the next-token entropy exceeds a threshold \citep[i.e., the model is uncertain;][]{wang2026beyond}, we crop the most-attended region of the cleaned map from the original image and insert it into the trace. Across 10 models V$^\ast$ \citep{wu2024vstar}, HR-Bench (4K and 8K) \citep{wang2024hrbench}, and MME-RealWorld-Lite \citep{zhang2025mme}, our method improves accuracy on 39 with a mean gain of $+4.3$ points. Because the key edit is cheap and leaves the model's outputs essentially unchanged, the cleaned map could replace the raw one wherever attention guides a VLM, from interpreting where the model looks to providing annotation-free grounding signals in RL pipelines.

\section{Related Work}
\label{sec: related work}

\subsection{Attention sinks}

Attention sinks are the few tokens that absorb much of the attention regardless of content \citep{xiao2024streamingllm}. They carry massive activations \citep{sun2024massive}, and their key acts as a bias while their value is near zero \citep{gu2025sink}. Vision transformers show the same on high-norm patches \citep{darcet2024registers}. A VLM inherits the text sink from its language model, and in some families image tokens behave like one as well. \citet{kang2025see} find these visual attention sinks through massive activations in a few hidden dimensions. Later work attributes them to rotary decay \citep{zhang2026drives}, separates encoder-born from language-model-born sinks \citep{choi2026sinks}, or reads them as carriers of semantics \citep{luo2026sink}. 

Prior work treats the attention on a visual sink as misplaced and redistributes it over the image \citep{kang2025see} or intervenes on it against hallucination \citep{huang2024opera,liu2024pai}. We find otherwise: in the full-attention models, removing a sink returns its attention to the text sink, not to the target, and the answers stay nearly the same. What changes is the attention map, which matters to every method that reads one, from attention-guided cropping to token pruning and grounding \citep{zhang2025vicrop,chen2024fastv,kang2025localization}.

\subsection{Visual Tools in Multi-modal Reasoning}

Attention to the image fades as the trace grows \citep{sun2025mitigating}, and a self-reflective phrase in the trace does not make the model look again \citep{shi2026vlms}. A phrase injected from outside helps only over the tokens already in context \citep{bi2026think}. Other methods act on the image. Visual programming composes vision experts and code around a language model \citep{gupta2023visprog,suris2023vipergpt}, Visual Sketchpad adds a drawing canvas \citep{hu2024sketchpad}, and a line of work trains a crop or zoom action into the trace \citep{zheng2025deepeyes,su2025pixelreasoner,zhang2025cof}. Training-free methods pick the region from the model's confidence \citep{shen2025zoomeye} or attention \citep{zhang2025vicrop,gao2025interleaved}. Only Look Twice \citep{morini2026look} filters sink tokens first with the score of \citet{kang2025see}.

Our re-look differs in where, what, and when. It reads the attention map after the key edit removes the sinks. It inserts a fresh crop as a second image block rather than encoded tokens \citep{gao2025interleaved} or a phrase \citep{bi2026think}. And it fires at the first uncertain step of the trace, since a few high-entropy tokens set its direction \citep{wang2026beyond}.

\begin{figure}[!b]
\centering
\vspace{-1em}
\includegraphics[width=\linewidth]{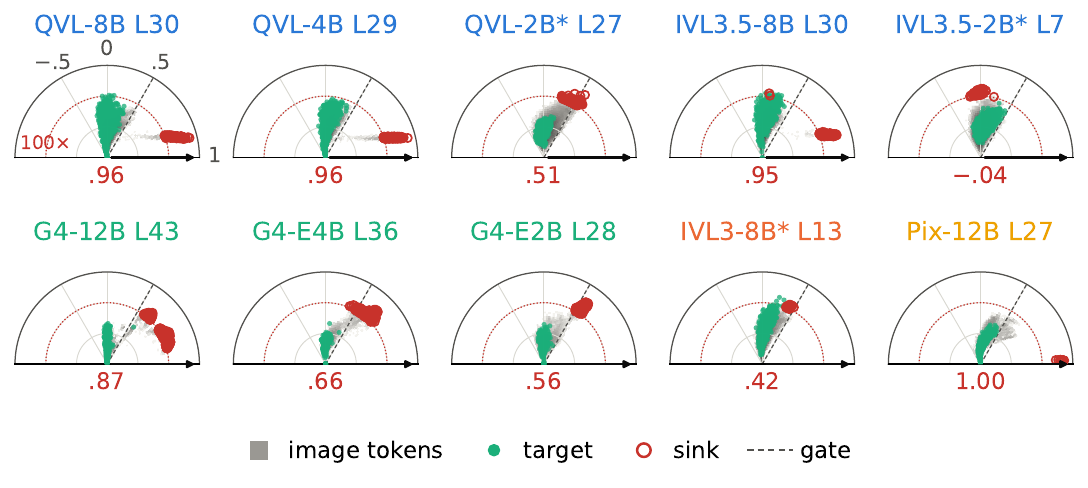}
\vspace{-2.5em}
\caption{Image tokens on V$^\ast$ at each model's peak layer. Angle: centered cosine of a token's key with the \tsink key (\autoref{eq:centered-cos}). Radius: attention received relative to uniform, log scale. Sinks: tokens above $100\times$ uniform (dotted ring) outside the target box. Dashed ray: gate threshold $.5$. Red numbers: median cosine of the sinks; $^\ast$: sink-free}
\label{fig:sink-keyalign}
\vspace{-1em}
\end{figure}

\section{Attention Sinks in VLMs Are Borrowed, Not Seen}
\label{sec:sink}

\providecommand{\usink}{\hat{u}}

We first identify which VLMs have \vsinks (\S~\ref{sec:sink-where}) and show that a sink is an image token whose key aligns with the \tsink key (\S~\ref{sec:sink-sig}). We then remove the sinks and measure where their attention goes and how the model's output changes (\S~\ref{sec:keyclean}).

All analyses use ten models from the Qwen3-VL \citep[QVL;][]{qwen2025qwen3vl}, InternVL \citep[IVL;][]{zhu2025internvl3,wang2025internvl35}, Gemma-4 \citep[G4][]{team2026gemma}, and Pixtral \citep[Pix][]{agrawal2024pixtral} families on V$^\ast$ \citep{wu2024vstar}, whose target boxes let us separate sinks from the queried object. HR-Bench 4K and 8K \citep{wang2024hrbench} have no target boxes but give matching results (\S~\ref{app:sink-hr}).

% This section presents a series of analyses organized around three questions: (1) What kind of VLMs show attention sinks? (2) What characterizes a \vsink? (3) What happens when a \vsink is removed? We answer those question in this section.

% \dengjia{we have mentioned those statemen in intro. im not sure whether we should mention it again here.}
% \vsink is \textbf{borrowed} because its key carries the key direction of the \tsink, and \textbf{not seen} because removing it leaves the model's next-token distribution essentially unchanged.

% All analyses use the V$^\ast$ dataset and models from the Qwen, Gemma, Pixtral, and InternVL families. We also provide results on HR-4K and HR-8K in \autoref{app:sink-hr}.

\subsection{Definitions}
A prompt ($\mathcal{IQ}$) is the $n$ image tokens $\mathcal{I}$ and its text question tokens $\mathcal{Q}$. For layer $\ell$ and head $h$, $\alpha^{\ell,h}_{i,j}$ is the prefill softmax attention from token $i$ to $j$. We measure the attention each token receives from the question and each image token's share of in-image attention:
\begin{equation}
a^{\ell}_{j} = \frac{1}{H|\mathcal{Q}|}\sum_{h}\sum_{i\in\mathcal{Q}}
\alpha^{\ell,h}_{i,j},
\qquad
r^{\ell}_{j} = \frac{a^{\ell}_{j}}{\sum_{j'\in\mathcal{I}} a^{\ell}_{j'}}
\;\; (j\in\mathcal{I}).
\label{eq:sink-notation}
\end{equation}
Uniform attention gives $r^\ell_j = 1/n$, so we report a token's attention relative to uniform ($n\,r^\ell_j$). 

Every model except Gemma-4 is full-attention. Most Gemma-4 layers instead use a sliding window.

\paragraph{Sinks.}
The \tsink $t^\star$ is the text position that receives the most attention from the question, $t^\star = \arg\max_{j\notin\mathcal{I}} a^{\ell}_j$, at the middle full-attention layer. A \vsink at layer $\ell$ is an image token that receives at least $100\times$ its uniform share, $n\,r^\ell_j \ge 100$, and lies outside the target bounding box $\mathcal{B}$ of the queried object.

\paragraph{Key directions.}
The key of token $j$ at layer $\ell$ and key--value head $g$ is $k^{\ell,g}_j$, the mean image key is $\bar k^{\ell,g} = \frac{1}{n} \sum_{j\in\mathcal{I}} k^{\ell,g}_j$, and the direction of the \tsink key is $\usink^{\ell,g} = k^{\ell,g}_{t^\star}/ \|k^{\ell,g}_{t^\star}\|$.

\paragraph{Sink-bearing and free models.}
The \vsink share $s^\ell = \sum_{j\in\mathcal{V}^\ell} r^\ell_j$ of layer $\ell$ is the in-image attention held by its \vsinks $\mathcal{V}^\ell$, averaged over images. A model's peak layer $\ell^\star$ is the layer with the largest $s^\ell$ in the last three quarters of its language model.\footnote{The first quarter is excluded because target localization is still weak there (\autoref{app:sink-analysis}).} A model is sink-bearing if $s^{\ell^\star} > 0.15$ and sink-free otherwise. We define this threshold only for analysis as a sink-free model under this definition may still have a few weak \vsinks. The key edit experiments in \S~\ref{sec:keyclean} treats every model alike, removing little from sink-free ones.

\subsection{Visual attention sinks come from the language model}
\label{sec:sink-where}

In \autoref{tab:sink-families}, we report the number of \vsinks per image, their attention relative to uniform, and their share of the in-image attention at each model's peak layer.\footnote{\autoref{tab:sink-families-full} gives all ten models and adds the text-sink and localization columns.} We find that seven of the ten models are sink-bearing, with sinks holding a quarter to four fifths of the in-image attention, and three are sink-free, with at most $12\%$.

\begin{wraptable}{r}{0.55\textwidth}
\vspace{-2em}
\centering
\caption{Visual attention sinks in VLMs on V$^\ast$ at each model's peak layer. key1: sinks per image; key 2: their attention relative to uniform; key 3: their share of the in-image attention.}
\label{tab:sink-families}
\vspace{-1em}
\setlength{\tabcolsep}{3pt}
\begin{tabular}{ll|ccc}
\toprule
Model & LM & per image & $\times$uniform & share \\
\midrule
QVL-8B    & Qwen3   & 3.9 & 1259 & .66 \\
QVL-2B    & Qwen3   & 2.7 &  163 & .12 \\
IVL3.5-8B & Qwen3   & 1.5 &  342 & .24 \\
IVL3.5-2B & Qwen3   & 1.0 &  151 & .08 \\
IVL3-8B   & Qwen2.5 & 0.4 &   93 & .02 \\
Pix-12B   & Mistral & 1.0 & 2093 & .76 \\
G4-12B    & Gemma-4 & 4.4 &  205 & .63 \\
G4-E2B    & Gemma-4 & 2.6 &  488 & .81 \\
\bottomrule
\end{tabular}
\vspace{-1em}
\end{wraptable}

The language model largely decides which group a model falls in. The Qwen3, Mistral, and Gemma-4 backbones produce sinks, while the Qwen2.5 backbone of InternVL3-8B does not. The clearest case is InternVL3-8B and InternVL3.5-8B, which share a vision encoder but use Qwen2.5 and Qwen3 respectively. With these models \vsinks take $2\%$ of the in-image attention in the first and $24\%$ in the second. Within a family, smaller models have weaker sinks. \Vsinks take $66\%$ and $73\%$ of the in-image attention in Qwen3-VL-8B and 4B but only $12\%$ in Qwen3-VL-2B, and $24\%$ in InternVL3.5-8B but $8\%$ in InternVL3.5-2B. 

Gemma-4 is sink-bearing for a different reason. Most of its layers attend only within a sliding window and the question sits too far after the \tsink to reach it. In these layers a few image tokens take $63\%$ to $81\%$ of the in-image attention. In its full-attention layers, which do reach the \tsink, no image token takes a large share.

\subsection{A sink is an image key aligned with the text-sink key}
\label{sec:sink-sig}

\begin{figure}[!t]
\centering
\vspace{-1em}
\includegraphics[width=\linewidth]{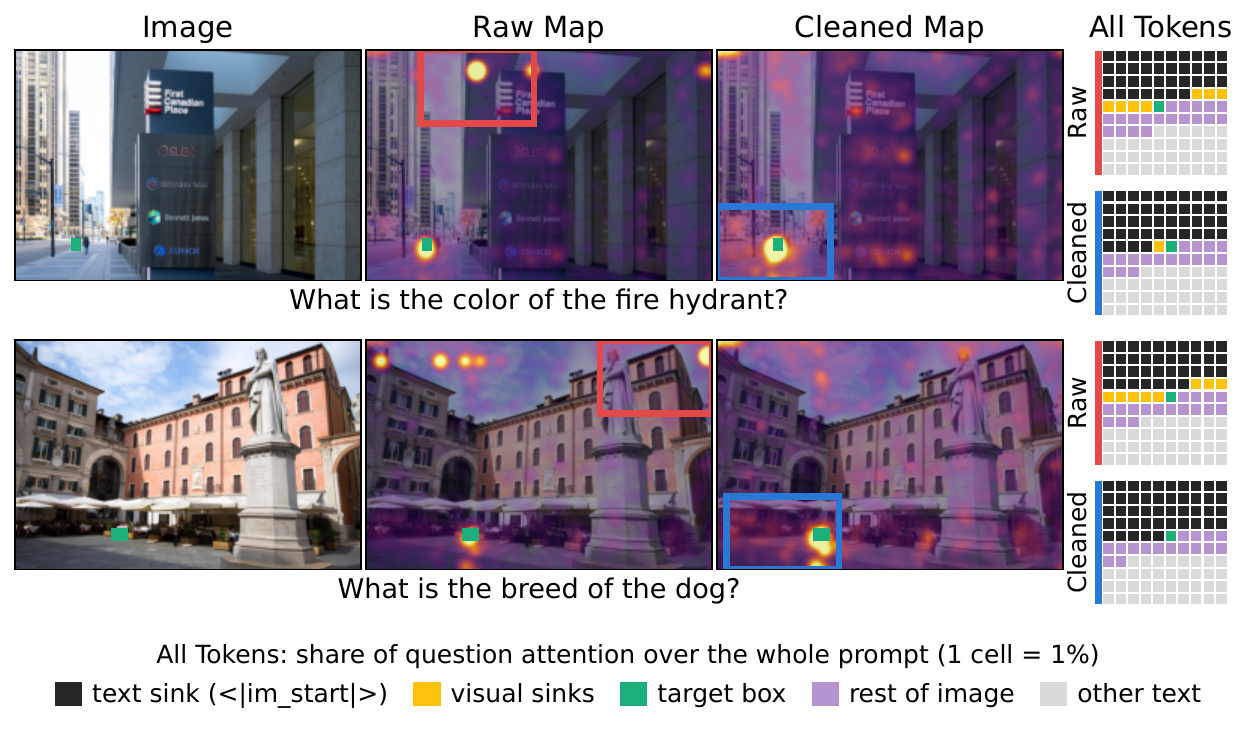}
\vspace{-2.5em}
\caption{Crop selection in Qwen3-VL-8B-Thinking on two V$^\ast$ images. Green: target box; red/blue: raw/edited window. Right: question attention over the whole prompt (one cell per $1\%$). The edit moves the \vsinks' share to the \tsink.}
\label{fig:sink-examples}
\vspace{-1.5em}
\end{figure}

Prior work identifies sinks by the size of their activations. We test three such features, the hidden-state norm \citep{kang2025see} and the value and key norms \citep{sun2024massive}, as detectors of \vsinks at each model's peak layer. We propose a detector based instead on the direction of a token's key. The \tsink acts as a bias that attracts every query regardless of content \citep{gu2025sink}, so if image tokens become sinks by the same mechanism, their keys should point along the \tsink key. We measure this with the centered cosine similarity between an image token's key and the \tsink key, which subtracts the mean image key $\bar k$ to remove the offset all image keys share
\begin{equation}
\tilde c_j=\frac{\langle k_j-\bar k,\;k_{t^\star}-\bar k\rangle}
{\|k_j-\bar k\|\,\|k_{t^\star}-\bar k\|},
\qquad
\tilde c_j^{\,\ell}=\frac{1}{H_{\mathrm{kv}}}\sum_{g}\tilde c_j^{\,\ell,g},
\label{eq:centered-cos}
\end{equation}
where the first expression is computed per layer and the key--value head, and the second averages it over the $H_{\mathrm{kv}}$ key--value heads. We also test two simpler versions of the same idea, the raw cosine without centering and the logit spread, the standard deviation of $\langle q_i,k_j\rangle/\sqrt{d}$ over question tokens $i$.

\paragraph{Detecting sinks.}
In \autoref{tab:sink-detectors-summary}, we report each feature's AUC at separating \vsinks from other image tokens, as the mean and minimum over the seven sink-bearing models. We find that only the centered cosine separates sinks in every model, with a minimum AUC of $99.0$, while the features from prior work are family-specific. Hidden-state and value norms separate sinks in the Qwen3 backbones and Pixtral but not reliably in Gemma-4, and key norm works only with its sign reversed, since sinks have unusually small keys, and still fails on Gemma-4. Of the other two new features, logit spread works only on Pixtral, and the raw cosine matches the centered one except on Gemma-4-E2B, where it drops to $94.0$. \autoref{fig:sink-keyalign} shows this separation as a compass. In every sink-bearing model the sinks are the only tokens that both point toward the \tsink key and receive high attention: their median cosine is $.95$ to $1.00$ in the full-attention models and $.56$ to $.87$ in Gemma-4, while the target and the other image tokens stay near cosine $0$ and receive far less attention.

\paragraph{Editing keys.}
To test whether the alignment causes sink attention, we edit keys at the peak layer of the four sink-bearing full-attention models on 20 V$^\ast$ images.\footnote{Full results are in \autoref{tab:sink-causal-ext}. For Pixtral, whose image keys as a whole sit at an offset along $\usink$, we move the sink keys' component to the median image key instead of zero (\S~\ref{app:sink-causal}).} First, we remove the direction. Projecting $\usink$ out of the sink keys \emph{drops their share of in-image attention to zero}, while projecting out a random direction leaves it unchanged. Second, we write the direction into other tokens. Setting the keys of eight random image tokens to the \tsink key \emph{gives them 91\% to 97\% of the in-image attention}, while a random key of the same norm gives them at most 9\%.

\begin{table}[t]
\centering
\caption{Effects of the key edit on V$^\ast$. $L$: number of language-model layers. \tsink: share of question attention on the \tsink, averaged over the probed layers}
\vspace{-0.5em}
\label{tab:sink-removal}
\setlength{\tabcolsep}{5pt}
\begin{tabular}{lccc|cccc}
\toprule
& \multicolumn{3}{c|}{Map} & \multicolumn{4}{c}{Model} \\
Model & $\max_j r_j$ & top-1 $\in\mathcal{B}$ & \tsink & accuracy & agreement & KL & caption \\
\midrule
Qwen3-VL-8B & $.25\to.04$ & $.00\to.47$ & $.57\to.67$ & $.73\to.73$ & .99 & .004 & .83 \\
Qwen3-VL-4B & $.24\to.05$ & $.00\to.41$ & $.54\to.68$ & $.73\to.73$ & .98 & .003 & .82 \\
Qwen3-VL-2B & $.04\to.03$ & $.28\to.23$ & $.63\to.63$ & $.68\to.61$ & .81 & .039 & .57 \\
InternVL3.5-8B & $.34\to.05$ & $.00\to.72$ & $.44\to.65$ & $.71\to.68$ & .94 & .001 & .85 \\
InternVL3.5-2B & $.04\to.04$ & $.60\to.59$ & $.65\to.65$ & $.64\to.61$ & .93 & .002 & .80 \\
InternVL3-8B & gate off & .59 & $.50\to.50$ & $.70\to.71$ & .97 & .001 & .87 \\
Pixtral-12B & $.57\to.02$ & $.00\to.16$ & $.40\to.51$ & $.50\to.50$ & .97 & .002 & .73 \\
\midrule
Gemma-4-12B & $.15\to.13$ & $.01\to.00$ & -- & $.68\to.65$ & .85 & .531 & .73 \\
Gemma-4-E4B & $.32\to.30$ & $.00\to.00$ & -- & $.44\to.44$ & .96 & .031 & .70 \\
Gemma-4-E2B & $.41\to.39$ & $.00\to.00$ & -- & $.39\to.36$ & .78 & .132 & .66 \\
\bottomrule
\end{tabular}
\vspace{-1em}
\end{table}

Alignment with the \tsink key is thus the signature of a \vsink. It separates sinks in every sink-bearing model, and in the models we edit it is both necessary and sufficient for sink attention. Because this direction belongs to the language model, the signature is also consistent with the backbone deciding which models have sinks (\S~\ref{sec:sink-where}).

\begin{wraptable}{r}{0.3\textwidth}
\vspace{0.5em}
\centering
\caption{AUC of each feature at separating \vsinks from other
image tokens, at the peak layer on V$^\ast$. Mean and min over sink-bearing models.}
\label{tab:sink-detectors-summary}
\setlength{\tabcolsep}{3pt}
\vspace{-0.5em}
\begin{tabular}{lcc}
\toprule
Feature & mean & min \\
\midrule
hidden norm & 80.1 & 25.0 \\
$-\|v\|$ & 72.0 & 8.0 \\
$-\|k\|$ & 77.4 & 49.0 \\
\midrule
$-$logit std       & 78.4 & 65.0 \\
raw key cos        & 99.0 & 94.0 \\
centered key cos   & 99.9 & 99.0 \\
\bottomrule
\end{tabular}
\vspace{-1.5em}
\end{wraptable}

\paragraph{Gemma-4.}
Gemma-4's sinks carry the same signature but play a different role. Shrinking the image until the whole prompt fits inside the sliding window moves a third to half of the question's attention in these layers back to the text, and attention on the strongest image token falls by two thirds or more (\S~\ref{app:sink-gemma-window}). Its image sinks therefore stand in for a \tsink the model cannot reach. Unlike the sinks of the full-attention models, they carry values of ordinary size and dominate the attention output of their layers, so the model does read them (\S~\ref{app:sink-value}).

\subsection{Removing the sinks leaves the model unchanged}
\label{sec:keyclean}

\paragraph{Removing \vsinks.}
Since the signature is a single direction, a sink can be removed by editing its key along that direction. At each layer and key--value head, an image token passes the gate when its centered cosine exceeds a threshold $\tau_c$ (\autoref{eq:centered-cos}). If more than a fraction $\rho$ of the layer's image tokens pass, the alignment belongs to the image as a whole rather than to a few sinks, and we leave the layer unedited. For each token that passes, we move its component along $\usink$ to the median over image keys,
\begin{equation}
k_j \leftarrow k_j-\bigl(\langle k_j,\usink\rangle
-\operatorname{median}_{j'\in\mathcal{I}}\langle k_{j'},\usink\rangle
\bigr)\,\usink ,
\label{eq:keyedit}
\end{equation}
with $\tau_c=0.5$ and $\rho=0.02$ for all models. In the full-attention models, any $\tau_c$ from $0.4$ to $0.9$ gives the same edit (\S~\ref{app:sink-heads}). The target is the median rather than zero because the image keys as a whole can sit at an offset along $\usink$, as in Pixtral, where a key with zero component would still stand out (\S~\ref{app:sink-causal}). Queries and values are untouched, so a gated token keeps its content and loses only its pull through the sink direction.

\paragraph{Where the attention goes.}
In \autoref{tab:sink-removal}, we report the effect of removing the sinks on the attention map and on the output for all ten models.\footnote{HR-Bench gives the same results (\autoref{tab:sink-removal-hr}).} We find that the attention the sinks held is borrowed from the \tsink. In the four sink-bearing full-attention models, the edit raises the \tsink's share of question attention by $0.10$ to $0.21$, while attention on the target box stays where it was. Forcing the freed attention onto the image instead raises target attention by up to $2.6\times$ but leaves accuracy unchanged, and it disturbs the output more than the key edit does (\S~\ref{app:sink-redistribution}).

\paragraph{Effect on the output.}
The model does not see the sinks. In the same four models, accuracy shifts by at most three points and first-token KL from the unedited model is of order $10^{-3}$ nats. The reason is that sinks pass on almost nothing. At the peak layer they hold $29\%$ to $77\%$ of the in-image attention but supply only $3\%$ to $16\%$ of the attention output norm, since their values are small and copy the \tsink's value direction (\S~\ref{app:sink-value}). Gemma-4 is the exception. As \S~\ref{sec:sink-sig} showed, the model reads its image sinks, and removing them changes the output (KL up to $0.53$) without clearing the map, so we do not apply the edit to Gemma-4. On the sink-free InternVL3 the gate admits under one token per image and the edit is an identity. On the two 2B sink-free models, the gate admits weakly aligned tokens and costs up to seven points, and restricting the edit to the upper three quarters of layers restores baseline accuracy.

\paragraph{Effect on the map.}
The attention map changes sharply. In the four sink-bearing full-attention models, the largest single-token share of in-image attention falls by an order of magnitude, and the most-attended image token, which never lay in the target box before the edit, lands in it on $16\%$ to $72\%$ of images afterwards (\autoref{fig:sink-examples}). The target gains no attention. It becomes the most-attended image token because the sinks no longer outrank it.

\section{Method}
\label{sec:method}

Section~\ref{sec:sink} showed that removing \vsinks leaves the model's answers unchanged but changes its attention map. The harm sinks do is therefore to anything that reads the map from outside the model. We study this harm through a task where reading the map correctly matters. On fine-detail questions, the evidence is often a small region of a large image, which the model may see only at reduced resolution. As the model reasons toward an answer, it reaches points of high uncertainty \citep{wang2026beyond}, and zooming into the relevant region is known to improve accuracy \citep{zhang2025vicrop, shen2025zoomeye, zheng2025deepeyes}. Finding that region is a map-reading problem, and \S~\ref{sec:keyclean} showed that the cleaned map places its most-attended token on the target far more often than the raw map.

\begin{figure}[!t]
    \centering
    \includegraphics[width=\linewidth]{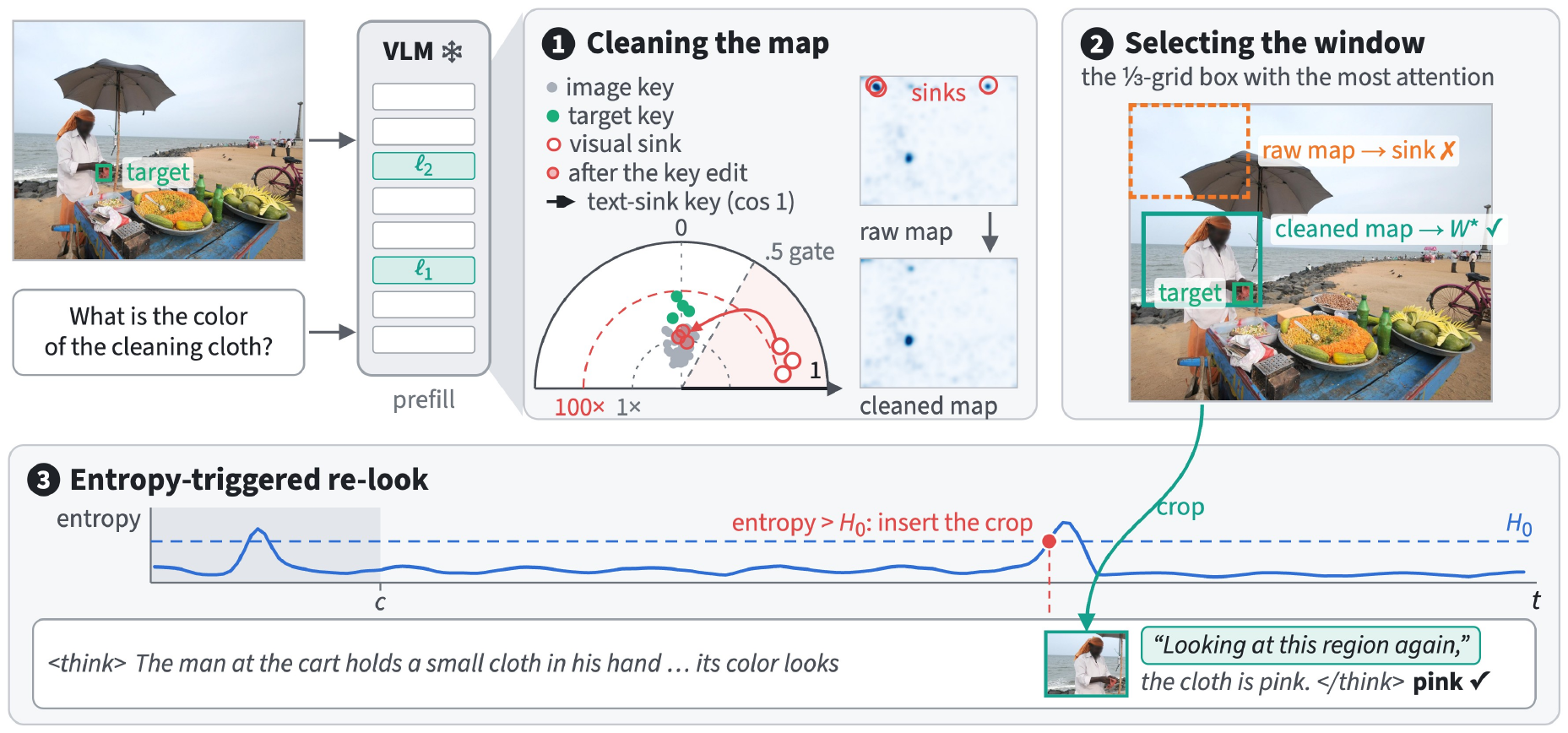}
    \caption{Overview of the entropy-triggered re-look. (1) The key edit removes the \vsinks from the prefill attention map. (2) The selector keeps the $\frac{1}{3}$-grid window with the most attention on the cleaned map. (3) When the next-token entropy first exceeds $H_0$ after $c$ tokens of thought, the crop of that window is inserted into the chain of thought.}
    \label{fig:method}
    \vspace{-1em}
\end{figure}

We combine the two in a training-free, entropy-triggered re-look that zooms into the region the cleaned map selects at the moment the model becomes uncertain (\autoref{fig:method}). First, the key edit of \S~\ref{sec:keyclean} removes the \vsinks from the attention map. The edit is applied only in a separate prefill that reads the map, so the model that writes the answer is unedited. Second, a selector reads the cleaned map and picks the image window with the most attention (\S~\ref{sec:selector}). Third, during chain-of-thought decoding, a trigger waits for the first step where the next-token entropy exceeds a threshold, then crops the selected window from the original image and inserts it into the trace (\S~\ref{sec:trigger}). Running the same re-look on the raw map isolates the effect of the sinks.

\subsection{Selecting the window}
\label{sec:selector}

The selector reads the question-to-image attention $a^\ell_j$ of \autoref{eq:sink-notation} from the edited prefill, averaged over two layers near half depth where the map responds most to the question, $\bar a_j = (a^{\ell_1}_j + a^{\ell_2}_j)/2$. Because Gemma-4 reads its image sinks, its selector instead uses two full-attention layers, which reach the \tsink and carry almost no image sinks, and skips the edit.

The window’s length is one-third of the side of the image-token grid. For tiled inputs, the tiles are first stitched back together into one grid. From the set of all same-sized windows $\mathcal{W}$ that lie within the grid, the selector chooses the window with the highest aggregate attention by applying a softmax over the region: $W^\star=\operatorname*{arg\,max}_{W\in\mathcal{W}}\sum_{j\in W}\bar a_j .$
% \begin{equation}
% W^\star=\operatorname*{arg\,max}_{W\in\mathcal{W}}\sum_{j\in W}\bar a_j .
% \label{eq:window-selection}
% \end{equation}
The window is mapped back to pixel coordinates, padded by 20 pixels on each side, and kept fixed for the rest of the question.

% The selector reads the head-averaged attention from question tokens to image tokens in the prefill stage  as a heatmap over the image. We take it at two middle layers where grounding is strongest and average them: $\bar a_j=(a_j^{\ell_1}+a_j^{\ell_2})/2$. The layer different model uses are shown in \autoref{app:selector-layers}.

% The window is the box that holds the most heat. We fix the box to one third of the token grid on each side, slide it over the grid, and keep the position with the largest sum:
% \begin{equation}
% W^\star=\operatorname*{arg\,max}_{W\in\mathcal{W}}\sum_{j\in W}\bar a_j ,
% \label{eq:window-selection}
% \end{equation}
% where $\mathcal{W}$ is the set of all boxes. The winning box is mapped back to pixel coordinates, padded by 20 pixels, and stays fixed for the rest of the question.

% The important thing is to clean the map first. On the sink-bearing models the \vsinks are already present at the selector layers and dominate the raw map, so without the key edit the box locates the sink rather than the object (\autoref{sec:keyclean}). Gemma-4 is the special case. Its sliding-window layers grow a different kind of sink, so we read two full-attention layers and skip the edit (\autoref{sec:exp}).

\subsection{Triggering and inserting the crop}
\label{sec:trigger}

The trigger decides when to insert the crop into the chain of thought. At every decoding step, it computes the entropy of the next-token distribution over the full vocabulary, after temperature scaling and before any top-$k$ or top-$p$ truncation. The crop is inserted at the first step where this entropy exceeds $H_0$, provided at least $c$ tokens of thought have been generated and decoding is still inside the thought channel. We use $H_0=1$ nat and $c=32$. The trigger fires at most once per question, and the token sampled at that step is kept.

At that step, we cut the selected window from the original pixels, so the crop is a zoom rather than a thumbnail, and encode it with the model's own image processor under the same budget cap as the first image. We append it to the reasoning trace as an ordinary second image, with no special positions or embeddings, followed by the text ``Looking at this region again,'' and decoding continues with the same sampling state. An untriggered question decodes exactly as the baseline.

Every question costs one extra prefill of the original prompt for the selector, and each triggered question adds one image encoding and one image-token block to the context. The weights are unchanged, and the crop and its transition text are the only additions to the prompt.

\section{Experiments}
\label{sec:exp}

\subsection{Setup}
\label{sec:exp:setup}

We run the re-look on the ten models of \autoref{sec:sink} with frozen weights. Eight reason in a native thinking mode, and we prompt the other two, the instruction models InternVL3-8B and Pixtral-12B, to produce a chain of thought. We evaluate on V$^\ast$ and HR-Bench 4K and 8K, where the evidence is a small patch of a large image, and on MME-RealWorld-Lite \citep{zhang2025mme}, which has 1{,}919 five-choice questions on high-resolution photographs, remote-sensing and monitoring imagery, diagrams, and driving scenes. Baseline and re-look runs share the sampling seed, so their traces are identical up to the trigger and every difference after it comes from the crop. To isolate the effect of the sinks, we repeat each run with the window selected on the raw map. We test each (model, benchmark) pair with an exact McNemar test. Prompts and sampling details are in \autoref{app:exp-details}.

\begin{table}[t]
  \caption{Base accuracy (\%) and paired re-look gain. \emph{Clean $\Delta$}: crop selected on the sink-cleaned map. \emph{Raw $\Delta$}: on the raw attention map. Shading scales with gain (green) or loss (red); $^{\dagger}$: $p<.05$ (exact McNemar, clean)}
  \vspace{-1em}
  \label{tab:main}
  \begin{center}
  \setlength{\tabcolsep}{3pt}
  \resizebox{\textwidth}{!}{%
  \begin{tabular}{l ccc ccc ccc ccc}
    \toprule
    & \multicolumn{3}{c}{\textbf{V$^{*}$}} & \multicolumn{3}{c}{\textbf{HR-4K}} & \multicolumn{3}{c}{\textbf{HR-8K}} & \multicolumn{3}{c}{\textbf{MME-RW-Lite}} \\
    \cmidrule(lr){2-4} \cmidrule(lr){5-7} \cmidrule(lr){8-10} \cmidrule(lr){11-13}
    \textbf{Model} & Base & Raw $\Delta$ & Clean $\Delta$ & Base & Raw $\Delta$ & Clean $\Delta$ & Base & Raw $\Delta$ & Clean $\Delta$ & Base & Raw $\Delta$ & Clean $\Delta$ \\
    \midrule
    \multicolumn{13}{l}{\textit{Qwen3-VL-Thinking}} \\
    \quad 8B & 79.1 & \cellcolor{green!15}$+2.1$ & \cellcolor{green!26}$+3.7$ & 72.6 & \cellcolor{green!5}$+0.8$ & \cellcolor{green!29}$+4.1$$^{\dagger}$ & 68.9 & \cellcolor{green!14}$+2.0$ & \cellcolor{green!31}$+4.4$$^{\dagger}$ & 46.5 & \cellcolor{green!7}$+1.0$ & \cellcolor{green!14}$+2.0$$^{\dagger}$ \\
    \quad 4B & 82.2 & \cellcolor{red!15}$-2.1$ & \cellcolor{green!18}$+2.6$ & 74.1 & \cellcolor{red!4}$-0.5$ & \cellcolor{green!6}$+0.9$ & 67.4 & \cellcolor{green!10}$+1.5$ & \cellcolor{green!36}$+5.1$$^{\dagger}$ & 45.1 & \cellcolor{green!5}$+0.8$ & \cellcolor{green!24}$+3.4$$^{\dagger}$ \\
    \quad 2B & 70.2 & \cellcolor{green!22}$+3.1$ & \cellcolor{green!11}$+1.6$ & 69.5 & \cellcolor{green!21}$+3.0$ & \cellcolor{green!21}$+3.0$$^{\dagger}$ & 63.9 & \cellcolor{green!29}$+4.1$ & \cellcolor{green!22}$+3.1$$^{\dagger}$ & 47.2 & \cellcolor{green!15}$+2.1$ & \cellcolor{green!14}$+2.0$$^{\dagger}$ \\
    \addlinespace[2pt]
    \multicolumn{13}{l}{\textit{InternVL3.5 (thinking)}} \\
    \quad 8B & 69.1 & \cellcolor{green!51}$+7.3$ & \cellcolor{green!62}$+8.9$$^{\dagger}$ & 64.5 & \cellcolor{green!44}$+6.2$ & \cellcolor{green!43}$+6.1$$^{\dagger}$ & 59.0 & \cellcolor{green!55}$+7.8$ & \cellcolor{green!49}$+7.0$$^{\dagger}$ & 48.7 & \cellcolor{green!15}$+2.1$ & \cellcolor{green!22}$+3.1$$^{\dagger}$ \\
    \quad 2B & 58.6 & \cellcolor{green!70}$+9.9$ & \cellcolor{green!73}$+10.5$$^{\dagger}$ & 62.4 & \cellcolor{green!32}$+4.6$ & \cellcolor{green!30}$+4.2$$^{\dagger}$ & 51.0 & \cellcolor{green!64}$+9.1$ & \cellcolor{green!66}$+9.5$$^{\dagger}$ & 44.2 & \cellcolor{green!1}$+0.1$ & \cellcolor{green!3}$+0.4$ \\
    \addlinespace[2pt]
    \multicolumn{13}{l}{\textit{Gemma-4 (thinking; selector reads the raw map)}} \\
    \quad 12B & 75.9 & -- & \cellcolor{green!48}$+6.8$$^{\dagger}$ & 76.4 & -- & \cellcolor{green!32}$+4.6$$^{\dagger}$ & 72.0 & -- & \cellcolor{green!69}$+9.9$$^{\dagger}$ & 52.5 & -- & \cellcolor{green!19}$+2.8$$^{\dagger}$ \\
    \quad E4B & 55.0 & -- & \cellcolor{green!75}$+11.0$$^{\dagger}$ & 68.1 & -- & \cellcolor{green!13}$+1.9$ & 62.4 & -- & \cellcolor{green!35}$+5.0$$^{\dagger}$ & 46.4 & -- & \cellcolor{green!26}$+3.6$$^{\dagger}$ \\
    \quad E2B & 50.8 & -- & \cellcolor{green!41}$+5.8$ & 66.3 & -- & \cellcolor{green!9}$+1.3$ & 57.9 & -- & \cellcolor{green!46}$+6.6$$^{\dagger}$ & 44.8 & -- & \cellcolor{red!3}$-0.5$ \\
    \addlinespace[2pt]
    \multicolumn{13}{l}{\textit{Chain-of-thought prompted}} \\
    \quad InternVL3-8B & 70.7 & \cellcolor{green!40}$+5.8$ & \cellcolor{green!40}$+5.8$ & 70.9 & \cellcolor{green!2}$+0.2$ & \cellcolor{green!3}$+0.4$ & 64.2 & \cellcolor{green!38}$+5.4$ & \cellcolor{green!38}$+5.4$$^{\dagger}$ & 48.8 & \cellcolor{green!7}$+0.9$ & \cellcolor{green!7}$+0.9$ \\
    \quad Pixtral-12B & 48.7 & \cellcolor{green!29}$+4.2$ & \cellcolor{green!51}$+7.3$ & 55.1 & \cellcolor{red!15}$-2.1$ & \cellcolor{green!8}$+1.1$ & 47.0 & \cellcolor{green!4}$+0.6$ & \cellcolor{green!22}$+3.1$ & 41.0 & \cellcolor{green!9}$+1.4$ & \cellcolor{green!26}$+3.7$$^{\dagger}$ \\
    \midrule
    Mean (all) & 66.0 & -- & \cellcolor{green!45}$+6.4$ & 68.0 & -- & \cellcolor{green!20}$+2.8$ & 61.4 & -- & \cellcolor{green!41}$+5.9$ & 46.5 & -- & \cellcolor{green!15}$+2.1$ \\
    Mean (with raw run) & 68.4 & \cellcolor{green!30}$+4.3$ & \cellcolor{green!41}$+5.8$ & 67.0 & \cellcolor{green!12}$+1.7$ & \cellcolor{green!20}$+2.8$ & 60.2 & \cellcolor{green!31}$+4.4$ & \cellcolor{green!38}$+5.4$ & 45.9 & \cellcolor{green!8}$+1.2$ & \cellcolor{green!15}$+2.2$ \\
    \bottomrule
  \end{tabular}}
  \end{center}
  \vspace{-1em}
\end{table}

\subsection{Results}
\label{sec:exp:results}

\paragraph{Overall.}
In \autoref{tab:main}, we report the baseline accuracy and the paired gain from one re-look for all 40 (model, benchmark) pairs. The re-look improves accuracy on all but one pair, by $4.3$ points on average. The gains concentrate on the fine-detail benchmarks, where the evidence is a small patch of a large image, averaging $5.0$ points on V$^\ast$ and HR-Bench against $2.1$ on MME-RealWorld-Lite, whose questions cover a wider range of tasks.

\paragraph{Raw versus cleaned map.}
The Raw columns of \autoref{tab:main} repeat each run with the window selected on the raw map. Gemma-4 has no raw run, since its selector already skips the edit (\S~\ref{sec:selector}). Over the 28 pairs with both runs, the cleaned map gains $4.1$ points on average and the raw map $2.9$, and the whole gap comes from the models with sinks. In Qwen3-VL-8B, Qwen3-VL-4B, and Pixtral-12B, the raw map gives back one to five points per pair, five pairs become significantly worse, and two, Qwen3-VL-4B on V$^\ast$ and Pixtral-12B on HR-4K, turn from gains into losses. The difference is where the crop lands. On Qwen3-VL-8B's V$^\ast$ questions, the raw crop misses the target in $78\%$ of triggers and the cleaned crop in $10\%$. InternVL3.5-8B, whose sinks are the weakest of the sink-bearing models, sits in between. On the sink-free models, the two maps pick nearly the same crops and the gains match, with InternVL3-8B differing on at most four questions per set. The same split holds on MME-RealWorld-Lite alone, at $2.2$ points cleaned against $1.2$ raw on average. Cleaning matters where there is a sink to clean.

\paragraph{Image resolution.}
The re-look helps most when the first pass could not see the evidence clearly. On V$^\ast$ and HR-Bench, Qwen3-VL, which has the largest image budget of the four families, gains about $3$ points on average, against about $7$ for InternVL3.5 and Gemma-4-12B. Within Qwen3-VL, all three sizes gain significantly on HR-8K, where the first pass sees about a third of the pixels, and none does on V$^\ast$, where the image already fits the budget. On a large image, the crop recovers pixels the first pass discarded. On an image the model already saw in full, it offers only a second look at the region, which is worth a few points at most.

\paragraph{Trigger coverage.}
An untriggered question decodes identically in both runs, so every gain comes from the triggered subset. On V$^\ast$ and HR-Bench, the trigger fires on $60\%$ to $78\%$ of questions for Qwen3-VL, $79\%$ or more for InternVL3.5, $91\%$ to $95\%$ for Pixtral-12B, and $96\%$ or more for InternVL3-8B and Gemma-4. On MME-RealWorld-Lite, it fires on $91\%$ or more for every model.

\section{Conclusion}
\label{sec:conclusion}

A \vsink is an image token whose key is aligned with the key of the language model's \tsink. This is the one feature that separates sinks from other image tokens in every family we test. Removing the direction from an image key removes the sink and writing it onto an image key creates one. The sink is inherited from the \tsink of the language model. When we remove it the attention flows back to the \tsink and the model's output does not change. Thus cleaning the attention map is a single cheap key edit. Building on this we design an entropy-triggered re-look. When a high-entropy token appears in the chain of thought the method crops the most-attended region of the cleaned map and inserts it into the trace. This improves accuracy where the same crop method from the raw map does not. Since cleaning the map costs almost nothing the method can also be used inside training pipelines to provide a grounding signal at no additional cost.

\section*{AI disclosure}
In this work, we used generative AI tools for two purposes: (1) polishing the writing, (2) retrieval and discovery of related work. We did not use generative AI tools for research ideation or research execution, and the remaining disclosure categories are not applicable to this work.

We use AIs for
\begin{itemize}
    \item Improve the grammar, wording, and clarity of author-written text. All content, claims, and technical details were written by the authors and every AI-edited passage was reviewed to ensure that the original meaning was preserved.
    \item AI-assisted search tools were used to supplement our manual literature survey and reduce the risk of overlooking relevant prior work. Every reference surfaced this way was independently located, read, and verified by the authors before being cited. 
\end{itemize}

We have reviewed all AI-assisted work and take full responsibility for the final content of this work, including all text, claims, and artifacts produced with the aid of generative AI.

\bibliography{iclr2027_conference}

@inproceedings{xiao2024streamingllm,
  title={Efficient streaming language models with attention sinks},
  author={Xiao, Guangxuan and Tian, Yuandong and Chen, Beidi and Han, Song and Lewis, Mike},
  booktitle={International Conference on Learning Representations},
  volume={2024},
  pages={21875--21895},
  year={2024}
}

@article{sun2024massive,
  title={Massive activations in large language models},
  author={Sun, Mingjie and Chen, Xinlei and Kolter, J Zico and Liu, Zhuang},
  journal={arXiv preprint arXiv:2402.17762},
  year={2024}
}

@inproceedings{gu2025sink,
  title={When attention sink emerges in language models: An empirical view},
  author={Gu, Xiangming and Pang, Tianyu and Du, Chao and Liu, Qian and Zhang, Fengzhuo and Du, Cunxiao and Wang, Ye and Lin, Min},
  booktitle={International Conference on Learning Representations},
  volume={2025},
  pages={97114--97144},
  year={2025}
}

@inproceedings{darcet2024registers,
  title={Vision transformers need registers},
  author={Darcet, Timoth{\'e}e and Oquab, Maxime and Mairal, Julien and Bojanowski, Piotr},
  booktitle={International conference on learning representations},
  volume={2024},
  pages={2632--2652},
  year={2024}
}

@inproceedings{kang2025see,
  title={See what you are told: Visual attention sink in large multimodal models},
  author={Kang, Seil and Kim, Jinyeong and Kim, Junhyeok and Hwang, Seong Jae},
  booktitle={International Conference on Learning Representations},
  volume={2025},
  pages={87676--87703},
  year={2025}
}

@article{barbero2025first,
  title={Why do {LLMs} attend to the first token?},
  author={Barbero, Federico and Arroyo, Alvaro and Gu, Xiangming and Perivolaropoulos, Christos and Bronstein, Michael and Veli{\v{c}}kovi{\'c}, Petar and Pascanu, Razvan},
  journal={arXiv preprint arXiv:2504.02732},
  year={2025}
}

@article{zhang2026drives,
  title={What drives attention sinks? a study of massive activations and rotational positional encoding in large vision--language models},
  author={Zhang, Xiaofeng and Zhu, Yuanchao and Gu, Chaochen and Cao, Jiawei and Cheng, Hao and Wu, Kaijie},
  journal={Information Processing \& Management},
  volume={63},
  number={2},
  pages={104431},
  year={2026},
  publisher={Elsevier}
}

@inproceedings{huang2024opera,
  title={{OPERA}: Alleviating hallucination in multi-modal large language models via over-trust penalty and retrospection-allocation},
  author={Huang, Qidong and Dong, Xiaoyi and Zhang, Pan and Wang, Bin and He, Conghui and Wang, Jiaqi and Lin, Dahua and Zhang, Weiming and Yu, Nenghai},
  booktitle={2024 IEEE/CVF Conference on Computer Vision and Pattern Recognition (CVPR)},
  pages={13418--13427},
  year={2024},
  organization={IEEE}
}

@inproceedings{liu2024pai,
  title={Paying more attention to image: A training-free method for alleviating hallucination in {LVLMs}},
  author={Liu, Shi and Zheng, Kecheng and Chen, Wei},
  booktitle={European Conference on Computer Vision},
  pages={125--140},
  year={2024},
  organization={Springer}
}

@inproceedings{chen2024fastv,
  title={An image is worth 1/2 tokens after layer 2: Plug-and-play inference acceleration for large vision-language models},
  author={Chen, Liang and Zhao, Haozhe and Liu, Tianyu and Bai, Shuai and Lin, Junyang and Zhou, Chang and Chang, Baobao},
  booktitle={European Conference on Computer Vision},
  pages={19--35},
  year={2024},
  organization={Springer}
}

@misc{zhang2025sparsevlm,
      title={SparseVLM: Visual Token Sparsification for Efficient Vision-Language Model Inference}, 
      author={Yuan Zhang and Chun-Kai Fan and Junpeng Ma and Wenzhao Zheng and Tao Huang and Kuan Cheng and Denis Gudovskiy and Tomoyuki Okuno and Yohei Nakata and Kurt Keutzer and Shanghang Zhang},
      year={2025},
      eprint={2410.04417},
      archivePrefix={arXiv},
      primaryClass={cs.CV},
      url={https://arxiv.org/abs/2410.04417}, 
}

@inproceedings{yang2025visionzip,
  title={{VisionZip}: Longer is better but not necessary in vision language models},
  author={Yang, Senqiao and Chen, Yukang and Tian, Zhuotao and Wang, Chengyao and Li, Jingyao and Yu, Bei and Jia, Jiaya},
  booktitle={2025 IEEE/CVF Conference on Computer Vision and Pattern Recognition (CVPR)},
  pages={19792--19802},
  year={2025},
  organization={IEEE}
}

@misc{qwen2025qwen3vl,
      title={Qwen3-VL Technical Report}, 
      author={Shuai Bai and Yuxuan Cai and Ruizhe Chen and Keqin Chen and Xionghui Chen and Zesen Cheng and Lianghao Deng and Wei Ding and Chang Gao and Chunjiang Ge and Wenbin Ge and Zhifang Guo and Qidong Huang and Jie Huang and Fei Huang and Binyuan Hui and Shutong Jiang and Zhaohai Li and Mingsheng Li and Mei Li and Kaixin Li and Zicheng Lin and Junyang Lin and Xuejing Liu and Jiawei Liu and Chenglong Liu and Yang Liu and Dayiheng Liu and Shixuan Liu and Dunjie Lu and Ruilin Luo and Chenxu Lv and Rui Men and Lingchen Meng and Xuancheng Ren and Xingzhang Ren and Sibo Song and Yuchong Sun and Jun Tang and Jianhong Tu and Jianqiang Wan and Peng Wang and Pengfei Wang and Qiuyue Wang and Yuxuan Wang and Tianbao Xie and Yiheng Xu and Haiyang Xu and Jin Xu and Zhibo Yang and Mingkun Yang and Jianxin Yang and An Yang and Bowen Yu and Fei Zhang and Hang Zhang and Xi Zhang and Bo Zheng and Humen Zhong and Jingren Zhou and Fan Zhou and Jing Zhou and Yuanzhi Zhu and Ke Zhu},
      year={2025},
      eprint={2511.21631},
      archivePrefix={arXiv},
      primaryClass={cs.CV},
      url={https://arxiv.org/abs/2511.21631}, 
}

@misc{wang2025internvl35,
      title={InternVL3.5: Advancing Open-Source Multimodal Models in Versatility, Reasoning, and Efficiency}, 
      author={Weiyun Wang and Zhangwei Gao and Lixin Gu and Hengjun Pu and Long Cui and Xingguang Wei and Zhaoyang Liu and Linglin Jing and Shenglong Ye and Jie Shao and Zhaokai Wang and Zhe Chen and Hongjie Zhang and Ganlin Yang and Haomin Wang and Qi Wei and Jinhui Yin and Wenhao Li and Erfei Cui and Guanzhou Chen and Zichen Ding and Changyao Tian and Zhenyu Wu and Jingjing Xie and Zehao Li and Bowen Yang and Yuchen Duan and Xuehui Wang and Zhi Hou and Haoran Hao and Tianyi Zhang and Songze Li and Xiangyu Zhao and Haodong Duan and Nianchen Deng and Bin Fu and Yinan He and Yi Wang and Conghui He and Botian Shi and Junjun He and Yingtong Xiong and Han Lv and Lijun Wu and Wenqi Shao and Kaipeng Zhang and Huipeng Deng and Biqing Qi and Jiaye Ge and Qipeng Guo and Wenwei Zhang and Songyang Zhang and Maosong Cao and Junyao Lin and Kexian Tang and Jianfei Gao and Haian Huang and Yuzhe Gu and Chengqi Lyu and Huanze Tang and Rui Wang and Haijun Lv and Wanli Ouyang and Limin Wang and Min Dou and Xizhou Zhu and Tong Lu and Dahua Lin and Jifeng Dai and Weijie Su and Bowen Zhou and Kai Chen and Yu Qiao and Wenhai Wang and Gen Luo},
      year={2025},
      eprint={2508.18265},
      archivePrefix={arXiv},
      primaryClass={cs.CV},
      url={https://arxiv.org/abs/2508.18265}, 
}

@misc{zhu2025internvl3,
      title={InternVL3: Exploring Advanced Training and Test-Time Recipes for Open-Source Multimodal Models}, 
      author={Jinguo Zhu and Weiyun Wang and Zhe Chen and Zhaoyang Liu and Shenglong Ye and Lixin Gu and Hao Tian and Yuchen Duan and Weijie Su and Jie Shao and Zhangwei Gao and Erfei Cui and Xuehui Wang and Yue Cao and Yangzhou Liu and Xingguang Wei and Hongjie Zhang and Haomin Wang and Weiye Xu and Hao Li and Jiahao Wang and Nianchen Deng and Songze Li and Yinan He and Tan Jiang and Jiapeng Luo and Yi Wang and Conghui He and Botian Shi and Xingcheng Zhang and Wenqi Shao and Junjun He and Yingtong Xiong and Wenwen Qu and Peng Sun and Penglong Jiao and Han Lv and Lijun Wu and Kaipeng Zhang and Huipeng Deng and Jiaye Ge and Kai Chen and Limin Wang and Min Dou and Lewei Lu and Xizhou Zhu and Tong Lu and Dahua Lin and Yu Qiao and Jifeng Dai and Wenhai Wang},
      year={2025},
      eprint={2504.10479},
      archivePrefix={arXiv},
      primaryClass={cs.CV},
      url={https://arxiv.org/abs/2504.10479}, 
}

@misc{agrawal2024pixtral,
      title={Pixtral 12B}, 
      author={Pravesh Agrawal and Szymon Antoniak and Emma Bou Hanna and Baptiste Bout and Devendra Chaplot and Jessica Chudnovsky and Diogo Costa and Baudouin De Monicault and Saurabh Garg and Theophile Gervet and Soham Ghosh and Amélie Héliou and Paul Jacob and Albert Q. Jiang and Kartik Khandelwal and Timothée Lacroix and Guillaume Lample and Diego Las Casas and Thibaut Lavril and Teven Le Scao and Andy Lo and William Marshall and Louis Martin and Arthur Mensch and Pavankumar Muddireddy and Valera Nemychnikova and Marie Pellat and Patrick Von Platen and Nikhil Raghuraman and Baptiste Rozière and Alexandre Sablayrolles and Lucile Saulnier and Romain Sauvestre and Wendy Shang and Roman Soletskyi and Lawrence Stewart and Pierre Stock and Joachim Studnia and Sandeep Subramanian and Sagar Vaze and Thomas Wang and Sophia Yang},
      year={2024},
      eprint={2410.07073},
      archivePrefix={arXiv},
      primaryClass={cs.CV},
      url={https://arxiv.org/abs/2410.07073}, 
}

@inproceedings{wu2024vstar,
  title={V*: Guided visual search as a core mechanism in multimodal {LLMs}},
  author={Wu, Penghao and Xie, Saining},
  booktitle={2024 IEEE/CVF Conference on Computer Vision and Pattern Recognition (CVPR)},
  pages={13084--13094},
  year={2024},
  organization={IEEE}
}

@inproceedings{wang2024hrbench,
  title={Divide, conquer and combine: A training-free framework for high-resolution image perception in multimodal large language models},
  author={Wang, Wenbin and Ding, Liang and Zeng, Minyan and Zhou, Xiabin and Shen, Li and Luo, Yong and Yu, Wei and Tao, Dacheng},
  booktitle={Proceedings of the AAAI Conference on Artificial Intelligence},
  volume={39},
  number={8},
  pages={7907--7915},
  year={2025}
}

@inproceedings{shen2025zoomeye,
    title = "{Z}oom{E}ye: Enhancing Multimodal {LLM}s with Human-Like Zooming Capabilities through Tree-Based Image Exploration",
    author = "Shen, Haozhan  and
      Zhao, Kangjia  and
      Zhao, Tiancheng  and
      Xu, Ruochen  and
      Zhang, Zilun  and
      Zhu, Mingwei  and
      Yin, Jianwei",
    editor = "Christodoulopoulos, Christos  and
      Chakraborty, Tanmoy  and
      Rose, Carolyn  and
      Peng, Violet",
    booktitle = "Proceedings of the 2025 Conference on Empirical Methods in Natural Language Processing",
    month = nov,
    year = "2025",
    address = "Suzhou, China",
    publisher = "Association for Computational Linguistics",
    url = "https://aclanthology.org/2025.emnlp-main.335/",
    doi = "10.18653/v1/2025.emnlp-main.335",
    pages = "6602--6618",
    ISBN = "979-8-89176-332-6"
}

@inproceedings{zheng2025deepeyes,
 author = {Zheng, Ziwei and Yang, Minghao and Hong, Jack and Zhao, Chenxiao and Xu, Guohai and Yang, Le and Shen, Chao and XingYu},
 booktitle = {International Conference on Learning Representations},
 editor = {C. Vondrick and B. Hariharan and C. Raffel and L. Pinto and D. Yang and A. Faust},
 pages = {126775--126798},
 title = {DeepEyes: Incentivizing "Thinking with Images" via Reinforcement Learning},
 url = {https://proceedings.iclr.cc/paper_files/paper/2026/file/cdb347d7516d52a9280ea9d6708911f6-Paper-Conference.pdf},
 volume = {2026},
 year = {2026}
}

@inproceedings{su2025pixelreasoner,
 author = {Su, Alex and Wang, Haozhe and Ren, Weiming and Lin, Fangzhen and Chen, Wenhu},
 booktitle = {Advances in Neural Information Processing Systems},
 doi = {10.52202/085713-0280},
 editor = {D. Belgrave and C. Zhang and H. Lin and R. Pascanu and P. Koniusz and M. Ghassemi and N. Chen},
 pages = {8222--8251},
 publisher = {Curran Associates, Inc.},
 title = {Pixel Reasoner: Incentivizing Pixel Space Reasoning via Curiosity-Driven Reinforcement Learning},
 url = {https://proceedings.neurips.cc/paper_files/paper/2025/file/0c38f54740062529aa4117a04b583f3c-Paper-Conference.pdf},
 volume = {38, Main Conference},
 year = {2025}
}

@misc{zhang2025cof,
      title={Adaptive Chain-of-Focus Reasoning via Dynamic Visual Search and Zooming for Efficient VLMs}, 
      author={Xintong Zhang and Zhi Gao and Bofei Zhang and Pengxiang Li and Xiaowen Zhang and Yang Liu and Tao Yuan and Yuwei Wu and Yunde Jia and Song-Chun Zhu and Qing Li},
      year={2025},
      eprint={2505.15436},
      archivePrefix={arXiv},
      primaryClass={cs.CV},
      url={https://arxiv.org/abs/2505.15436}, 
}

@inproceedings{lai2025minio3,
  title={{Mini-o3}: Scaling up reasoning patterns and interaction turns for visual search},
  author={Lai, Xin and Li, Junyi and Li, Wei and Liu, Tao and Li, Tianjian and Zhao, Hengshuang},
  booktitle={International Conference on Learning Representations},
  volume={2026},
  pages={76722--76746},
  year={2026}
}

@inproceedings{wang2026beyond,
 author = {Wang, Shenzhi and Yu, Le and Gao, Chang and Zheng, Chujie and Liu, Shixuan and Lu, Rui and Dang, Kai and Chen, Xiong-Hui and Yang, Jianxin and Zhang, Zhenru and Liu, Yuqiong and Yang, An and Zhao, Andrew and Yue, Yang and Song, Shiji and Yu, Bowen and Huang, Gao and Lin, Junyang},
 booktitle = {Advances in Neural Information Processing Systems},
 doi = {10.52202/085713-3850},
 editor = {D. Belgrave and C. Zhang and H. Lin and R. Pascanu and P. Koniusz and M. Ghassemi and N. Chen},
 pages = {115452--115486},
 publisher = {Curran Associates, Inc.},
 title = {Beyond the 80/20 Rule: High-Entropy Minority Tokens Drive Effective Reinforcement Learning for LLM Reasoning},
 url = {https://proceedings.neurips.cc/paper_files/paper/2025/file/a797c2d2e0c1fdabf4d1ab8cd0b465c6-Paper-Conference.pdf},
 volume = {38, Main Conference},
 year = {2025}
}

@inproceedings{gupta2023visprog,
  title={Visual programming: Compositional visual reasoning without training},
  author={Gupta, Tanmay and Kembhavi, Aniruddha},
  booktitle={2023 IEEE/CVF Conference on Computer Vision and Pattern Recognition (CVPR)},
  pages={14953--14962},
  year={2023},
  organization={IEEE}
}

@inproceedings{suris2023vipergpt,
  title={{ViperGPT}: Visual inference via python execution for reasoning},
  author={Sur{\'\i}s, D{\'\i}dac and Menon, Sachit and Vondrick, Carl},
  booktitle={2023 IEEE/CVF International Conference on Computer Vision (ICCV)},
  pages={11854--11864},
  year={2023},
  organization={IEEE}
}

@article{hu2024sketchpad,
  title={Visual sketchpad: Sketching as a visual chain of thought for multimodal language models},
  author={Hu, Yushi and Shi, Weijia and Fu, Xingyu and Roth, Dan and Ostendorf, Mari and Zettlemoyer, Luke and Smith, Noah A and Krishna, Ranjay},
  journal={Advances in Neural Information Processing Systems},
  volume={37},
  pages={139348--139379},
  year={2024}
}

@inproceedings{zhang2025vicrop,
  author       = {Jiarui Zhang and
                  Mahyar Khayatkhoei and
                  Prateek Chhikara and
                  Filip Ilievski},
  title        = {MLLMs Know Where to Look: Training-free Perception of Small Visual
                  Details with Multimodal LLMs},
  booktitle    = {The Thirteenth International Conference on Learning Representations,
                  {ICLR} 2025, Singapore, April 24-28, 2025},
  publisher    = {OpenReview.net},
  year         = {2025},
  url          = {https://openreview.net/forum?id=DgaY5mDdmT},
  bibsource    = {dblp computer science bibliography, https://dblp.org}
}

@inproceedings{zhang2025mme,
 author = {Zhang, YiFan and Zhang, Huanyu and Tian, Haochen and Fu, Chaoyou and Zhang, Shuangqing and Wu, Junfei and Li, Feng and Wang, Kun and Wen, Qingsong and Zhang, Zhang and Wang, Liang and Jin, Rong},
 booktitle = {International Conference on Learning Representations},
 editor = {Y. Yue and A. Garg and N. Peng and F. Sha and R. Yu},
 pages = {89655--89701},
 title = {MME-RealWorld: Could Your Multimodal LLM Challenge High-Resolution Real-World Scenarios that are Difficult for Humans?},
 url = {https://proceedings.iclr.cc/paper_files/paper/2025/file/df29d63af05cb91d705cf06ba5945b9d-Paper-Conference.pdf},
 volume = {2025},
 year = {2025}
}

@inproceedings{liu2026seeing,
 author = {Liu, Zhining and Chen, Ziyi and Liu, Hui and Luo, Chen and Tang, Xianfeng and Wang, Suhang and Zeng, Jingying and Dai, Zhenwei and Shi, Zhan and Wei, Tianxin and Lu, Hanqing and Dumoulin, Benoit and Tong, Hanghang},
 booktitle = {International Conference on Learning Representations},
 editor = {C. Vondrick and B. Hariharan and C. Raffel and L. Pinto and D. Yang and A. Faust},
 pages = {72849--72876},
 title = {Seeing but Not Believing: Probing the Disconnect Between Visual Attention and Answer Correctness in VLMs},
 url = {https://proceedings.iclr.cc/paper_files/paper/2026/file/76818d8d85e05e45ce3a16a8468619d1-Paper-Conference.pdf},
 volume = {2026},
 year = {2026}
}

@inproceedings{kang2025localization,
  title={Your large vision-language model only needs a few attention heads for visual grounding},
  author={Kang, Seil and Kim, Jinyeong and Kim, Junhyeok and Hwang, Seong Jae},
  booktitle={2025 IEEE/CVF Conference on Computer Vision and Pattern Recognition (CVPR)},
  pages={9339--9350},
  year={2025},
  organization={IEEE}
}

@misc{su2026attention,
      title={Attention Sink in Transformers: A Survey on Utilization, Interpretation, and Mitigation}, 
      author={Zunhai Su and Hengyuan Zhang and Wei Wu and Yifan Zhang and Yaxiu Liu and He Xiao and Qingyao Yang and Yuxuan Sun and Rui Yang and Chao Zhang and Jing Xiong and Hui Shen and Keyu Fan and Weihao Ye and Chaofan Tao and Taiqiang Wu and Zhongwei Wan and Tiantian Zhang and Bowen Yan and Zhen Li and Yiming Zhang and Congkai Xie and Yulei Qian and Yuchen Xie and Yik-Chung Wu and Hongxia Yang and Ngai Wong},
      year={2026},
      eprint={2604.10098},
      archivePrefix={arXiv},
      primaryClass={cs.LG},
      url={https://arxiv.org/abs/2604.10098}, 
}

@inproceedings{luo2026sink,
  title={To sink or not to sink: Visual information pathways in large vision-language models},
  author={Luo, Jiayun and Fan, Wan-Cyuan Chris and Wang, Lyuyang and He, Xiangteng and Rahman, Tanzila and Abolmaesumi, Purang and Sigal, Leonid},
  booktitle={International Conference on Learning Representations},
  volume={2026},
  pages={93626--93660},
  year={2026}
}

@inproceedings{choi2026sinks,
  title={When sinks help or hurt: Unified framework for attention sink in large vision-language models},
  author={Choi, Jiho and Kim, Jaemin and Kim, Sanghwan and Hong, Seunghoon and Park, Jin-Hwi},
  booktitle={European Conference on Computer Vision},
  pages={148--166},
  year={2026},
  organization={Springer}
}

@article{li2026sinkpruner,
  title={SinkPruner: Sink-Free Visual Token Pruning for Multimodal Large Language Models},
  author={Li, Shiyu and Hu, Zi-Yuan and Huang, Shijia and Li, Yanyang and Zhong, Yiwu and Wang, Liwei},
  journal={arXiv preprint arXiv:2609.01004},
  year={2026}
}

@article{shukla2026sage,
  title={SAGE: Sink-Aware Grounded Decoding for Multimodal Hallucination Mitigation},
  author={Shukla, Tripti and Kira, Zsolt},
  journal={arXiv preprint arXiv:2603.27898},
  year={2026}
}

@inproceedings{kaduri2025s,
  title={What’s in the Imageƒ A Deep-Dive into the Vision of Vision Language Models},
  author={Kaduri, Omri and Bagon, Shai and Dekel, Tali},
  booktitle={2025 IEEE/CVF Conference on Computer Vision and Pattern Recognition (CVPR)},
  pages={14549--14558},
  year={2025},
  organization={IEEE}
}

@inproceedings{neo2025towards,
  title={Towards interpreting visual information processing in vision-language models},
  author={Neo, Clement and Ong, Luke and Torr, Philip and Geva, Mor and Krueger, David and Barez, Fazl},
  booktitle={International Conference on Learning Representations},
  volume={2025},
  pages={57172--57189},
  year={2025}
}

@inproceedings{jiang2025devils,
  title={Devils in middle layers of large vision-language models: Interpreting, detecting and mitigating object hallucinations via attention lens},
  author={Jiang, Zhangqi and Chen, Junkai and Zhu, Beier and Luo, Tingjin and Shen, Yankun and Yang, Xu},
  booktitle={2025 IEEE/CVF Conference on Computer Vision and Pattern Recognition (CVPR)},
  pages={25004--25014},
  year={2025},
  organization={IEEE}
}

@inproceedings{che2025easy,
  title={Hallucinatory image tokens: A training-free eazy approach to detecting and mitigating object hallucinations in lvlms},
  author={Che, Liwei and Liu, Tony Qingze and Jia, Jing and Qin, Weiyi and Tang, Ruixiang and Pavlovic, Vladimir},
  booktitle={2025 IEEE/CVF International Conference on Computer Vision (ICCV)},
  pages={21635--21644},
  year={2025},
  organization={IEEE}
}

@misc{team2026gemma,
      title={Gemma 4 Technical Report}, 
      author={Gemma Team and Sherif El Abd and Vaibhav Aggarwal and Robin Algayres and Alek Andreev and Olivier Bachem and Ian Ballantyne and Cormac Brick and Victor Cărbune and Michelle Casbon and Mayank Chaturvedi and Aditya Chawla and Victor Cotruta and Alice Coucke and Phil Culliton and Robert Dadashi and Lucas Dixon and Mohamed Elhawaty and Utku Evci and Clément Farabet and Johan Ferret and Filippo Galgani and Sertan Girgin and Jean-Bastien Grill and Maarten Grootendorst and Jiaxian Guo and Cassidy Hardin and Yanzhang He and Steven M. Hernandez and Omri Homburger and Léonard Hussenot and Juyeong Ji and Armand Joulin and Aishwarya Kamath and Parnian Kassraie and Olivier Lacombe and Preethi Lahoti and Gaël Liu and Gus Martins and Luciano Martins and Tatiana Matejovicova and Ramona Merhej and Nikola Momchev and Sneha Mondal and Ryan Mullins and Sindhu Raghuram Panyam and Shreya Pathak and Sarah Perrin and André Susano Pinto and Etienne Pot and Angéline Pouget and Alexandre Ramé and Sabela Ramos and Douglas Reid and David Rim and Morgane Rivière and Karsten Roth and Louis Rouillard and Omar Sanseviero and Pier Giuseppe Sessa and Shane Settle and Danila Sinopalnikov and Sara Smoot and Piotr Stanczyk and Andreas Steiner and Lawrence Stewart and Ilya Tolstikhin and Michael Tschannen and Anton Tsitsulin and Nino Vieillard and Renjie Wu and Pingmei Xu and Haichuan Yang and Edouard Yvinec and Biao Zhang and Li Zhang and Joe Zou and Nicolas Aagnes and Abdelrahman Abdelhamed and Jakub Adamek and Shivani Agrawal and Shubham Agrawal and Ibrahim Alabdulmohsin and Jean Baptiste Alayrac and Uri Alon and Chandramouli Amarnath and Ankesh Anand and Chrysovalantis Anastasiou and Setareh Ariafar and François-Xavier Aubet and Kyriakos Axiotis and Federico Barbero and Joelle Barral and Alexei Bendebury and Urs Bergmann and Stanley Bileschi and Kat Black and Mathieu Blondel and Sebastian Borgeaud and Arthur Bražinskas and Ryan Burnell and Robert Busa-Fekete and Mu Cai and Daniele Calandriello and Glenn Cameron and Charlotte Caucheteux and Rahma Chaabouni and Garima Chadha and Jetha Chan and Blake Jianhang Chen and Jesse Chen and Lin Chen and Xu Chen and Derek Cheng and Tzu-hsiang Chien and Nikolai Chinaev and Yi Chou and Zhaohui Chu and Benjamin Coleman and Pooja Consul and Sam Conway-Rahman and Scott Crowell and Dylan Cutler and Vivek Dani and Samira Daruki and Anil Das and Daniel Deutsch and Nishanth Dikkala and Li Ding and Qiuhan Ding and Shenil Dodhia and Konstantin Donhauser and Tulsee Doshi and Anca Dragan and Alex Druinsky and Sahil Dua and Zoltan Egyed and Danielle Eisenbud and Daniel Eppens and Cindy Fan and Bahare Fatemi and Yassir Fathullah and Vlad Feinberg and Milen Ferev and Sebastian Flennerhag and Takumi Fujimoto and João Gabriel Oliveira and Isaac Galatzer-Levy and João Gante and Simon Geisler and Soham Ghosal and Antonious M. Girgis and Tamara von Glehn and Alec Go and Alhaad Gokhale and Alex Grills and Yiming Gu and Mayank Gupta and Pramod Gupta and Guru Guruganesh and Raia Hadsell and Hamza Harkous and Jitendra Harlalka and Demis Hassabis and Anja Hauth and Joe Heyward and Arian Hosseini and Chih-Yang Hsia and I-Hung Hsu and Xiaopeng Huang and Yangsibo Huang and Kevin Hui and Adrian Hutter and Te I and Fotis Iliopoulos and Advait Jain and Ganesh Jawahar and Ziwei Ji and Qilin Jin and Melvin Johnson and Kandarp Joshi and Arun Kandoor and Wang-Cheng Kang and Koray Kavukcuoglu and Mehran Kazemi and Kathleen Kenealy and Amr Khalifa and Phoebe Kirk and Ivan Korotkov and Suraj Kothawade and Vitaly Kovalev and Neel Kovelamudi and Adam Kraft and Ravin Kumar and Vivek Kumar and Harish Kuppam and Justin Lannin and Chen-Yu Lee and Seungji Lee and Dmitry Lepikhin and Alon Levkovitch and Dongdong Li and Qiujia Li and Valentin Liévin and Ethan Lin and Ziqian Lin and Casper Liu and Tianlin Liu and Tianqi Liu and Xin Liu and Ivan Lobov and Mayank Lunayach and Min Ma and Gagan Madan and Andrii Maksai and Eric Malmi and Michal Matuszak and Daniel McDuff and Gaurav Menghani and Maciej Mikuła and Daniil Mirylenka and Karolis Misiunas and Vedant Misra and Andreea Mitran and Kareem Mohamed and Maksim Mukha and Eric Noland and James O'Donnell and Brendan O'Donoghue and Kate Olszewska and Bernett Orlando and Wanqiong Pan and Rina Panigrahy and Unnati Parekh and Nicolas Perez-Nieves and Chunjong Park and Eric Paskie and Liqian Peng and Bryce Petrini and Slav Petrov and Jonas Pfeiffer and Bilal Piot and Martyna Plomecka and Siim Poder and Octavio Ponce and Arijit Pramanik and David Racz and Anish Rajan and Michelle Ramanovich and Anand Rao and Marvin Ritter and Vitor Rodrigues and Evan Rosen and Mikołaj Rybiński and Noveen Sachdeva and Michaël E. Sander and Rohit Sathyanarayana and Sagar Savla and Samuel Schmidgall and Tal Schuster and George Scrivener and Benoit Seguin and Andrew Sellergren and Aliaksei Severyn and Izhak Shafran and Dhruv Shah and Bobak Shahriari and Yuan Shangguan and Ashish Shenoy and Pradeep Shenoy and Rakesh Shivanna and Pauline Sho and Lucas Spangher and Wojciech Stokowiec and Tim Strother and Yao Su and Yinghao Sun and Mukund Sundararajan and Andrea Tacchetti and Mor Hazan Taege and Pouya Tafti and Jean Tarbouriech and Chetan Tekur and Shantanu Thakoor and Rahul Thapa and Madeleine Traverse and Lenart Treven and Tao Tu and Chien Te Tung and Çağlar Ünlü and Petar Veličković and Malini Pooni Venkat and Sagar Gubbi Venkatesh and Vidya Venkiteswaran and Francesco Visin and Alex Vitvitskyi and Kiran Vodrahalli and Weiyi Wang and Xin Wang and Tris Warkentin and Jan Wassenberg and John Wieting and Cindy Wu and Lechao Xiao and Hao Xu and Yuhui Xu and Fuzhao Xue and Arun Yadav and Jun Yan and Antoine Yang and Lin Yang and Ming-Hsuan Yang and Ziyu Ying and Jae Hyeon Yoo and Morteza Zadimoghaddam and Sajjad Zafar and Fred Zhang and Jiageng Zhang and Jianyi Zhang and Xiaofan Zhang and Chao Zhao and David Zhou and Chen Zou},
      year={2026},
      eprint={2607.02770},
      archivePrefix={arXiv},
      primaryClass={cs.CL},
      url={https://arxiv.org/abs/2607.02770}, 
}

@inproceedings{sun2025mitigating,
  title={Mitigating visual forgetting via take-along visual conditioning for multi-modal long cot reasoning},
  author={Sun, Hai-Long and Sun, Zhun and Peng, Houwen and Ye, Han-Jia},
  booktitle={Proceedings of the 63rd Annual Meeting of the Association for Computational Linguistics (Volume 1: Long Papers)},
  pages={5158--5171},
  year={2025}
}

@article{shi2026vlms,
  title={Are vlms seeing or just saying? uncovering the illusion of visual re-examination},
  author={Shi, Chufan and Yang, Cheng and Wu, Yaokang and Jin, Linghao and Shui, Bo and Berg-Kirkpatrick, Taylor and Ma, Xuezhe},
  journal={arXiv preprint arXiv:2605.15864},
  year={2026}
}

@InProceedings{bi2026think,
    author    = {Bi, Jing and Bellos, Filippos and Guo, Junjia and Li, Yayuan and Huang, Chao and Tang, Yunlong and Song, Luchuan and Liang, Susan and Zhang, Zhongfei and Corso, Jason J. and Xu, Chenliang},
    title     = {When to Think and When to Look: Uncertainty-Guided Lookback},
    booktitle = {Proceedings of the IEEE/CVF Conference on Computer Vision and Pattern Recognition (CVPR)},
    month     = {June},
    year      = {2026},
    pages     = {5104-5113}
}

@inproceedings{gao2025interleaved,
  title={Interleaved-modal chain-of-thought},
  author={Gao, Jun and Li, Yongqi and Cao, Ziqiang and Li, Wenjie},
  booktitle={2025 IEEE/CVF Conference on Computer Vision and Pattern Recognition (CVPR)},
  pages={19520--19529},
  year={2025},
  organization={IEEE}
}

@misc{morini2026look,
      title={Look Twice: Training-Free Evidence Highlighting for Knowledge-based Visual Question Answering}, 
      author={Marco Morini and Sara Sarto and Marcella Cornia and Lorenzo Baraldi and Rita Cucchiara},
      year={2026},
      eprint={2604.01280},
      archivePrefix={arXiv},
      primaryClass={cs.CV},
      url={https://arxiv.org/abs/2604.01280}, 
}
\bibliographystyle{iclr2027_conference}

\appendix

\section{Sink Measurements on V$^\ast$}\label{app:sink-analysis}

\subsection{Setup and the ten models}
\label{app:sink-distribution}

We measure every softmax attention layer and head on V$^\ast$. Its target boxes cover $0.3$ to $0.4\%$ of the image tokens, depending on the model's tokenization. Prompts and image budgets are those of \autoref{tab:main}. Under those budgets a V$^\ast$ image is about 3{,}300 image tokens for Qwen3-VL and an HR-Bench image about 12{,}400. InternVL3.5 runs under its R1 system prompt and Gemma-4 under its thinking template. Every number below depends on that prompt. The accuracies in \autoref{tab:sink-removal} and \autoref{tab:sink-removal-hr} are greedy and letter-only with an empty thinking block, so they are not comparable with the chain-of-thought accuracies of \autoref{tab:main}. \autoref{tab:sink-families-full} extends \autoref{tab:sink-families} of the main text with the text-sink position and share and with the localization columns.

\begin{table}[!htbp]
\centering
\caption{Text and visual sinks in ten models on V$^\ast$, extending \autoref{tab:sink-families}. LM is the language-model backbone. Text sink: the token index of the text sink and its share of the question's attention. Visual sink (peak layer): attention relative to uniform, count per image, share of the in-image attention and the peak layer out of the model's layer count. Target top-1$\in\mathcal{B}$: the fraction of images whose most-attended image token lies in the target box, at the peak layer and at the best layer.}
\label{tab:sink-families-full}
\footnotesize\setlength{\tabcolsep}{4.5pt}
\resizebox{\linewidth}{!}{%
\begin{tabular}{llcccccccc}
\toprule
& & \multicolumn{2}{c}{Text sink} & \multicolumn{4}{c}{Visual sink (peak layer)} & \multicolumn{2}{c}{Target top-1$\in\mathcal{B}$} \\
\cmidrule(lr){3-4}\cmidrule(lr){5-8}\cmidrule(lr){9-10}
Model & LM & position & share & $\times$uniform & per image & share & layer & at peak & best layer \\
\midrule
Qwen3-VL-8B & Qwen3 & 0 & .74 & 1259 & 3.9 & .66 & 30/36 & .00 & .80 \\
Qwen3-VL-4B & Qwen3 & 0 & .69 & 1049 & 4.9 & .73 & 29/36 & .00 & .82 \\
Qwen3-VL-2B & Qwen3 & 0 & .75 & 163 & 2.7 & .12 & 27/28 & .00 & .75 \\
InternVL3.5-8B & Qwen3 & 0 & .65 & 342 & 1.5 & .24 & 30/36 & .14 & .62 \\
InternVL3.5-2B & Qwen3 & 0 & .77 & 151 & 1.0 & .08 & 7/28 & .00 & .69 \\
Gemma-4-12B & Gemma-4 & 2 & .10 & 205 & 4.4 & .63 & 43/48 & .00 & .66 \\
Gemma-4-E4B & Gemma-4 & 2 & .21 & 444 & 2.7 & .66 & 36/42 & .00 & .53 \\
Gemma-4-E2B & Gemma-4 & 1 & .22 & 488 & 2.6 & .81 & 28/35 & .00 & .24 \\
InternVL3-8B & Qwen2.5 & 2 & .62 & 93 & 0.4 & .02 & 13/28 & .05 & .76 \\
Pixtral-12B & Mistral & 0 & .49 & 2093 & 1.0 & .76 & 27/40 & .00 & .00 \\
\bottomrule
\end{tabular}}
\end{table}

\paragraph{Text-sink position.}
The \tsink $t^\star$ of \S \ref{sec:sink} is fixed per model. In the full-attention models it is also the text position with the largest share of the question's attention at every layer from about a third of the depth onward. Which position it is depends only on the backbone. Qwen3 (in Qwen3-VL and InternVL3.5) and Mistral (in Pixtral-12B) use the first token, and Qwen2.5 (in InternVL3-8B) uses a newline in the system prompt. The share is $.62$ to $.77$ in the full-attention models, except Pixtral-12B at $.49$. Gemma-4 has only seven or eight full-attention layers. In those layers the sink is the turn marker just after \texttt{<bos>}, at position 2 in Gemma-4-12B and Gemma-4-E4B and at position 1 in Gemma-4-E2B (\autoref{tab:sink-families-full}). It takes $.10$ to $.22$ of the question's attention, a sixth to a third of what the other backbones give their sink.

\paragraph{Which models have visual sinks.}
At their peak layer Qwen3-VL-8B, Qwen3-VL-4B, InternVL3.5-8B, Pixtral-12B and the three Gemma-4 sizes have image tokens at $200$ to $2{,}100\times$ uniform attention. One to five such tokens take a quarter to four fifths of the in-image attention. InternVL3-8B has only a faint version ($93\times$ uniform, $2\%$ share). Qwen3-VL-2B's tokens sit just above $100\times$ uniform and hold only a $12\%$ share. In InternVL3.5-2B a single such token appears at layer 7. Its key lacks the signature of \S \ref{sec:sink-sig} (centered cosine $-.03$, AUC $.42$). It holds the same share on every dataset and is gone from layer 8 on. Layer 7 is the first layer the peak-layer rule admits. By the $15\%$ cut of \S \ref{sec:sink} this makes Qwen3-VL-8B, Qwen3-VL-4B, InternVL3.5-8B, Pixtral-12B and the three Gemma-4 sizes sink-bearing and Qwen3-VL-2B, InternVL3.5-2B and InternVL3-8B sink-free.

\paragraph{Query independence.}
Does the sink set depend on the question? We replace the V$^\ast$ question by a generic prompt and recompute the late-layer sink set. It is almost unchanged (Jaccard overlap $.73$ to $.92$).

\begin{figure}[t]
\centering
\includegraphics[width=\linewidth]{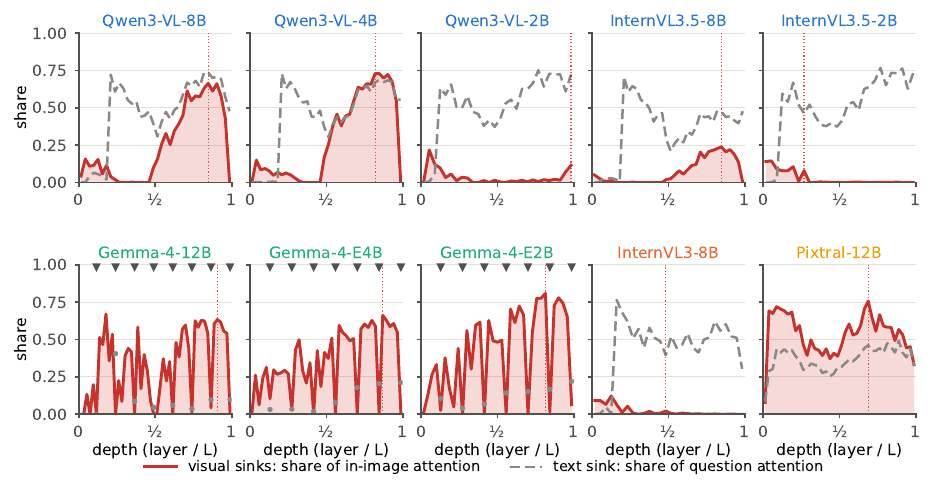}
\caption{Sink attention across depth on V$^\ast$ for all ten models. Red: \vsink share of in-image attention. Gray dashed: text-sink share of question attention. Dotted: peak layer. Gemma-4 reaches the text sink only in its full-attention layers (triangles). Title color marks the language-model backbone (Qwen3, Gemma-4, Qwen2.5 and Mistral).}
\label{fig:sink-depth-all}
\end{figure}

\paragraph{Sink depth.}
\autoref{fig:sink-depth-all} shows at which layers the \vsinks appear in each of the ten models. Qwen3 sinks switch on near half depth and persist to the penultimate layer. Pixtral-12B has one sink at a fixed grid position at every layer but the first few, and we call it the register. Gemma-4's sinks appear only in its sliding-window layers, where the \tsink is out of reach. In every model but Pixtral-12B the sinks collapse in the last layer. Pixtral-12B's register keeps a third of the in-image attention there.

\paragraph{Localization depth.}
The raw-map curves of \autoref{fig:sink-removal-grounding} track where the most-attended image token lands at each layer. In Qwen3-VL-8B localization peaks at mid depth and drops to zero once the sinks take over. InternVL3.5-8B keeps localizing while its sinks grow. On the raw map Pixtral-12B never localizes, and InternVL3-8B does from mid depth on. Gemma-4 localizes in its full-attention layers, the only Gemma-4 layers the figure plots. At its sliding-window peak layer the most-attended token is never in the box (\autoref{tab:sink-families-full}). At the peak layer localization is $.00$ in six of the seven sink-bearing models and $.14$ in InternVL3.5-8B (\autoref{tab:sink-families-full}). At the best layer five of the seven sink-bearing models reach $.53$ to $.82$ and Gemma-4-E2B reaches $.24$, while Pixtral-12B's register stays on top at every layer. This is why the selector reads mid-depth layers (\S \ref{app:selector-layers}).

\subsection{The first quarter of depth}
\label{app:sink-first-quarter}

Why does the peak-layer rule of \S \ref{sec:sink} skip the first quarter of each model's depth? Localization is weak there (the gray band of \autoref{fig:sink-removal-grounding}). In the first quarter the most-attended image token lies in the target box on $0$ to $9\%$ of V$^\ast$ images in every model. The rate first reaches $20\%$ between layers 9 and 17, and never in Pixtral-12B. Tokens above $100\times$ uniform do appear in the first quarter of every model, with or without later sinks.

Without the rule the peak layer would fall in the first quarter for four models. On V$^\ast$ Qwen3-VL-2B and InternVL3.5-2B would peak at layer 1, InternVL3-8B at layer 3 and Gemma-4-12B at layer 8. The same four models move on both HR-Bench sets. Gemma-4-12B is the one sink-bearing model among them. Its sliding-window layers carry sinks from layer 4 on, with a share comparable to the late layers. The gate of \S~\ref{sec:keyclean} acts per layer, and the selector reads mid-depth layers (\S~\ref{sec:selector}). The method is therefore unaffected either way.

The centered cosine of the sliding-window sinks stays below $.4$ in the first quarter of Gemma-4-12B, Gemma-4-E4B and Gemma-4-E2B. From mid depth on it settles at $.5$ to $.8$, the alignment reported in \S~\ref{sec:sink-sig}. Gemma-4's early sinks are therefore not yet aligned with the text-sink key. We have not tested whether the direction is carried up the residual stream from the full-attention layers. On Gemma-4-12B, the one size whose late-layer keys are its own (\S~\ref{app:sink-detectors}), the gate does not flag these early sinks. The selector never reads sliding-window layers on Gemma-4, so it never sees them.

\subsection{Detectors}
\label{app:sink-detectors}

\autoref{tab:sink-detectors} scores six token features as detectors of the attention-defined sinks at each model's peak layer. Each cell holds three numbers. Here $k$ is the number of sinks in the image. Precision at $k$ is the fraction of the feature's top-$k$ tokens that are sinks. The top-1 entry asks whether the feature's top token is the most-attended one. AUC measures the separation of sinks from all other tokens, and it can round to $1.00$ while a sink still ranks only in the top $1\%$ of three thousand tokens. We call a feature a detector when its AUC is at least $.99$ in every sink-bearing model. Precision at $k$ then asks whether it also names the individual sinks.

\begin{table}[t]
\centering
\caption{Token features as detectors of visual sinks at each model's peak layer on V$^\ast$. Cells: precision at $k$ / top-1 / AUC, where $k$ is the number of sinks in the image. The three sink-free models are below the rule. There the few tokens above $100\times$ uniform (0.4 to 2.7 per image) stand in for the sinks. The first two columns are the centered cosine between the token's key and the text-sink key (\autoref{eq:centered-cos}) and the same cosine without centering. The next five are the negated value norm, the hidden-state norm, the key norm with both signs and the negated standard deviation of the token's attention logits. Apart from the key norm, every feature is signed so that a larger value is more sink-like. Hidden norm is the feature of \citet{kang2025see}.}
\label{tab:sink-detectors}
\footnotesize\setlength{\tabcolsep}{3.5pt}
\resizebox{\linewidth}{!}{%
\begin{tabular}{lccccccc}
\toprule
Model & centered key cos & raw key cos & $-\|v\|$ & hidden norm & $\|k\|$ & $-\|k\|$ & $-$logit std \\
\midrule
Qwen3-VL-8B & .99/.98/1.00 & .99/.96/1.00 & .99/.94/1.00 & .99/.93/1.00 & .00/.00/.01 & .23/.00/.99 & .00/.00/.69 \\
Qwen3-VL-4B & .99/.95/1.00 & .99/.93/1.00 & .99/.68/1.00 & .98/.80/1.00 & .00/.00/.28 & .01/.00/.72 & .00/.00/.71 \\
InternVL3.5-8B & .95/.70/1.00 & .95/.66/1.00 & .95/.73/1.00 & .95/.61/.99 & .00/.00/.01 & .74/.28/.99 & .00/.00/.90 \\
Gemma-4-12B & .61/.00/1.00 & .61/.00/1.00 & .00/.00/.40 & .00/.00/.93 & .01/.00/.33 & .18/.21/.68 & .02/.00/.85 \\
Gemma-4-E4B & .18/.00/1.00 & .04/.00/.99 & .00/.00/.56 & .00/.00/.44 & .00/.00/.50 & .00/.00/.49 & .00/.00/.65 \\
Gemma-4-E2B & .07/.00/.99 & .00/.00/.94 & .00/.00/.08 & .00/.00/.25 & .06/.00/.46 & .15/.12/.55 & .00/.00/.69 \\
Pixtral-12B & 1.00/1.00/1.00 & 1.00/1.00/1.00 & 1.00/1.00/1.00 & .00/.00/1.00 & .00/.00/.00 & 1.00/1.00/1.00 & 1.00/1.00/1.00 \\
\midrule
\multicolumn{8}{l}{\emph{sink-free by the 15\% cut (\autoref{sec:sink}); the tokens above $100\times$ uniform stand in for the sinks}} \\
Qwen3-VL-2B & .51/.30/.99 & .78/.73/1.00 & .00/.00/.12 & .00/.00/.21 & .00/.00/.05 & .00/.00/.95 & .00/.00/.10 \\
InternVL3.5-2B & .00/.00/.42 & .01/.00/.91 & .00/.00/.75 & .00/.00/.06 & .00/.00/.87 & .00/.00/.13 & .00/.00/.48 \\
InternVL3-8B & .13/.09/1.00 & .50/.35/1.00 & .53/.45/1.00 & .00/.00/.30 & .00/.00/.24 & .00/.00/.76 & .00/.00/.30 \\
\bottomrule
\end{tabular}}
\end{table}

The centered key cosine has an AUC of at least $.99$ in every sink-bearing model, and in the full-attention families it also names the individual sinks. It is the one feature that meets the criterion. In Gemma-4 the AUC stays high but precision is low. Other tokens there sit even closer to the weak text-sink key and outrank the sinks. The raw key cosine matches the centered one in the full-attention families and on Gemma-4-12B. On Gemma-4-E4B and Gemma-4-E2B its precision falls to $.04$ and $.00$, and on Gemma-4-E2B the AUC falls to $.94$.

A low value norm finds the sinks in the Qwen3 backbones and Pixtral-12B. It does not find them in Gemma-4, whose sinks have values of ordinary size. The hidden-state norm is the massive-activation signature of \citet{kang2025see} and \citet{sun2024massive}. By the AUC criterion it passes in the Qwen3 backbones and Pixtral-12B and fails in Gemma-4. In the Qwen3 backbones it also names the individual sinks. In Pixtral-12B it does not, because the register is a massive activation but never the largest in the image and some border tokens outrank it (precision $.00$), just as other tokens outrank the Gemma-4 sinks under the centered cosine. In hidden norm the Gemma-4 sinks stay within a factor of two of the median at every depth. A massive activation is therefore one route to the sink direction. It is not the only one. The key norm itself has AUC near $0$ in Qwen3-VL-8B, InternVL3.5-8B and Pixtral-12B, so the sinks have the smallest keys there and the negated key norm separates them (AUC $.99$ to $1.00$). It names them in Pixtral-12B and partly in InternVL3.5-8B (precision $.74$) but hardly in Qwen3-VL-8B ($.23$). It fails in Qwen3-VL-4B (AUC $.72$) and in Gemma-4. The logit spread separates the sinks only in Pixtral-12B.

The rows of the three sink-free models describe 0.4 to 2.7 tokens per image above $100\times$ uniform. The centered cosine separates them in Qwen3-VL-2B and InternVL3-8B (AUC $.99$ and $1.00$) but not in InternVL3.5-2B ($.42$), whose single layer-7 token lacks the key signature (\S \ref{app:sink-distribution}).

Gemma-4-E2B and Gemma-4-E4B share keys and values across their last 20 and 18 layers (from layers 15 and 24), so their peak-layer keys are those of an earlier layer. Only Gemma-4-12B has its own keys at its peak layer. The key and value features in the Gemma-4-E4B and Gemma-4-E2B rows of \autoref{tab:sink-detectors} are therefore those of layers 24 and 15, scored against the attention of layers 36 and 28.

\subsection{Key interventions}
\label{app:sink-causal}

\autoref{tab:sink-causal-ext} gives the numbers behind the causal claim of \S \ref{sec:sink-sig}. We run two interventions on Qwen3-VL-8B, Qwen3-VL-4B, InternVL3.5-8B and Pixtral-12B at each model's peak layer on 20 V$^\ast$ images. \emph{Remove} projects $\usink$ out of the key of every image token whose cosine to $\usink$ exceeds $.5$. This cosine is uncentered, unlike the gate of \S \ref{sec:keyclean}, which is why the counts below differ from those of \S \ref{app:sink-heads}. The table lists those tokens' in-image share before and after. The control projects out a random unit direction instead. On these 20 images the gate admits five tokens per image on Qwen3-VL-8B, seven on Qwen3-VL-4B and one on Pixtral-12B. Fifteen of the 20 images have such a token on InternVL3.5-8B. \emph{Align} takes $k=8$ random image tokens and sets their keys to the text-sink key. The Align columns list their share before and after. The control sets their keys to a random key of the same norm.

\begin{table}[h]
\centering
\caption{Key interventions at the peak layer on 20 V$^\ast$ images. \emph{Remove}: in-image share of the gated tokens before and after projecting out $\usink$, with a random unit direction as the control. \emph{Align}: share of $k{=}8$ random image tokens before and after setting their keys to $k_{t^\star}$, with a random key of the same norm as the control. \emph{Median target}: the $\usink$-component of the edited keys, in the Remove edit and in the Align random-key control, is moved to the population median instead of zero.}
\label{tab:sink-causal-ext}
\footnotesize\setlength{\tabcolsep}{5pt}
\begin{tabular}{llccc ccc}
\toprule
& & \multicolumn{3}{c}{Remove} & \multicolumn{3}{c}{Align} \\
\cmidrule(lr){3-5}\cmidrule(lr){6-8}
Model & Layer & before & $-\usink$ & $-$random & before & $:=k_{t^\star}$ & random key \\
\midrule
Qwen3-VL-8B    & 30 & .68 & .00 & .66 & .00 & .97 & .04 \\
Qwen3-VL-4B    & 29 & .75 & .00 & .74 & .00 & .95 & .00 \\
InternVL3.5-8B & 30 & .36 & .00 & .35 & .00 & .96 & .09 \\
Pixtral-12B    & 27 & .49 & .25 & .49 & .00 & .91 & .64 \\
\quad median target & 27 & .49 & .00 & .49 & .00 & .91 & .01 \\ 
\bottomrule
\end{tabular}
\end{table}

On the three Qwen3 backbones projecting out $\usink$ removes the sinks' share entirely and the random direction leaves it alone (\autoref{tab:sink-causal-ext}). Random keys aligned with the text-sink key capture nearly all the in-image attention. The full key edit of \S \ref{sec:keyclean} behaves the same way at every layer. The removed mass returns to the \tsink (text-sink column of \autoref{tab:sink-removal}).

Pixtral-12B's image keys point away from the \tsink (centered cosine $-.59$ to $\usink$ for the population). A key with zero component along $\usink$ therefore sits far above the population. With the projection to zero, the register's share only halves ($.49$ to $.25$). The same projection applied to random non-sink tokens on the same images lifts their share from $.00$ to $.56$. It does nothing on Qwen3-VL-8B, whose image keys sit near zero along $\usink$. Moving the register's component to the population median instead removes it ($.49$ to $.00$), and the alignment control separates cleanly ($.91$ versus $.01$). The key edit of \S \ref{sec:keyclean} therefore takes the median as its target.

\subsection{Gemma-4: what the sliding window cannot reach}
\label{app:sink-gemma-window}

Are Gemma-4's image sinks a substitute for a \tsink that the sliding window cannot reach? The released budget is 1{,}120 soft tokens, of which about 1{,}080 are image tokens on V$^\ast$. At that budget the text before the image lies more than one window back, so the question tokens cannot attend to it in the sliding-window layers. \autoref{tab:sink-gemma-window} repeats the sink measurements with the budget cut to 280 soft tokens (261 image tokens). This puts the whole prompt inside the window. The run uses the same 100 V$^\ast$ images and the same prompt as the released-budget run.

At the released budget the image takes $.51$ to $.60$ of the question's attention in these layers. With the text in reach it takes $.14$ to $.19$, and $.37$ to $.49$ goes to the text before the image. The strongest image token's absolute attention drops three- to five-fold. Its share within the image changes little in two of the three models ($.37$ to $.35$ in Gemma-4-E2B and $.17$ to $.19$ in Gemma-4-12B). In Gemma-4-E4B it falls from $.30$ to $.18$. The attention spreads over the text before the image. The text-sink position itself takes little of it. The image tokens of these layers therefore hold attention that the text would absorb if it were reachable.

The count of tokens above $100\times$ uniform also falls. That threshold is not comparable across budgets, because $100\times$ uniform is a much larger share of the image at 261 tokens than at 1{,}080. The smaller budget coarsens the image as well, so two things change at once.

\begin{table}[h]
\centering
\caption{Gemma-4 sliding-window layers at 1{,}120 versus 280 soft tokens on 100 V$^\ast$ images, averaged over heads, layers and images. Each cell reads released budget $\to$ reduced budget. The \emph{text before image} and \emph{image} columns are shares of the question's attention. The strongest image token's share of the question's attention is \emph{top-1, absolute}, and its share of the in-image attention is \emph{top-1, share of image}.}
\label{tab:sink-gemma-window}
\footnotesize\setlength{\tabcolsep}{5pt}
\begin{tabular}{lcccc}
\toprule
Model & text before image & image & top-1, absolute & top-1, share of image \\
\midrule
Gemma-4-E2B & $.00\to.38$ & $.55\to.19$ & $.21\to.07$ & $.37\to.35$ \\
Gemma-4-E4B & $.00\to.37$ & $.51\to.16$ & $.16\to.03$ & $.30\to.18$ \\
Gemma-4-12B & $.00\to.49$ & $.60\to.14$ & $.10\to.03$ & $.17\to.19$ \\
\bottomrule
\end{tabular}
\end{table} %% job 670712 (slurm_key/sink_r2_gemma_window.slurm) -> /exp/dzhang1/scratch/sink_r2_ext/gemma_window/; scripts/sink/r2ext/gemma_window_compare.py -> compare.json

\subsection{Heads and the gate threshold}
\label{app:sink-heads}

\begin{figure}[t]
\centering
\includegraphics[width=\linewidth]{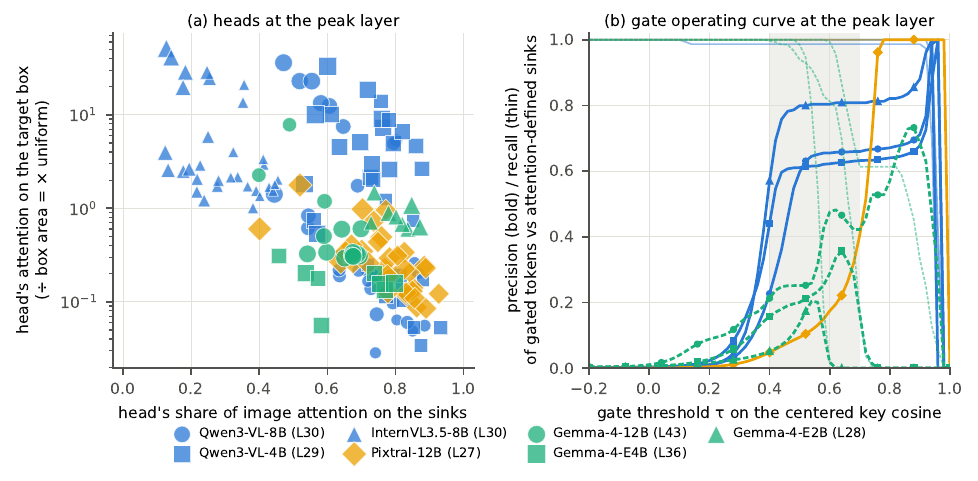}
\caption{(a) Sink share vs.\ target attention per head at the peak layer on V$^\ast$. Marker size is the head's share of image attention. (b) Precision and recall of the centered-cosine gate at threshold $\tau_c$ against the attention-defined sinks, one curve per sink-bearing model on V$^\ast$. Shaded band: thresholds that give the same edit. The paper uses 0.5.}
\label{fig:sink-heads}
\end{figure}

Does a head that sinks also localize? \autoref{fig:sink-heads}a plots each head's sink share against its attention on the target at the peak layer. In every sink-bearing model the two are negatively correlated. This is not mass conservation. The target box covers $0.3\%$ of a V$^\ast$ image. A head that gives $.8$ of its image attention to the sinks could still put its remaining $.2$ on the target at $50\times$ uniform. Such heads sit at $0.1\times$ instead, below uniform. Heads that sink under half their image attention reach $10$ to $30\times$. Different heads sink and localize, as different layers do.

How much does the edit depend on $\tau_c$? \autoref{fig:sink-heads}b plots the precision and recall of the gate against the attention-defined sinks (image tokens outside the target box above $100\times$ uniform, \autoref{sec:sink}) as $\tau_c$ varies. In the full-attention families recall is $1.0$ at every threshold from $0.4$ to $0.9$, so the gate never misses a sink. Precision is flat over the same range, at $.6$ on Qwen3-VL and $.8$ on InternVL3.5. Any threshold in the band gives the same edit. We use $0.5$. Below $0.4$ the gate admits ordinary tokens and precision collapses. On Qwen3-VL-8B it admits 17 per image at $0.4$ and 84 at $0.3$.

Which tokens does the gate admit inside the band? At $0.5$ the gate of \autoref{fig:sink-heads}b admits about seven tokens per image on Qwen3-VL-8B (five on the 20-image subset of \autoref{tab:sink-causal-ext}). Four of them are sinks by the $100\times$ definition. The other three sit at $10$ to $100\times$ uniform with keys aligned just as tightly. They are sinks below an arbitrary attention cut, and we want them removed too. Precision below $1$ inside the band is therefore not a false-positive problem.

Pixtral-12B and Gemma-4 fall outside this picture. In Pixtral-12B the band admits about ten tokens around its single register. Only $\tau_c>0.8$ isolates the register. The extra tokens do no harm, since the cleaned map still beats the raw one on Pixtral-12B (\autoref{tab:main}). Gemma-4 stays below $.5$ precision at every threshold and loses recall beyond $0.7$, which is a second reason, besides the output shift in \autoref{tab:sink-removal}, that the re-look selector skips the edit on Gemma-4 (\S~\ref{sec:selector}).

The cap $\rho=0.02$ is set by hand and not swept. It matters only below the band. At $\tau_c=0.3$ the 84 admitted tokens are $2.5\%$ of a 3{,}300-token image. That is above the cap, so the layer is left alone.

\subsection{What the sinks contribute to the output}
\label{app:sink-value}

Does the attention a sink receives reach the output? A large attention weight is not a large contribution. The output for query $i$ is $\sum_j\alpha_{ij}v_j$, so a token with a near-zero value adds little however much attention it gets. A value aligned with the \tsink's adds in a direction the \tsink already supplies. In the full-attention families a sink's value has both properties. Its norm is under a tenth of the image median and close to the \tsink's own value norm. The cosine between its direction and the \tsink's value direction is $.7$ to $.8$, against a population median of $-.1$. At each model's peak layer and in each head we measure the norm of the \vsinks' contribution relative to the full output. We also record the cosine between the output with and without that contribution (\autoref{tab:sink-value}).

\begin{table}[t]
\centering
\caption{Visual-sink contributions to the attention output at the peak layer on V$^\ast$. The \emph{attention} column is the sinks' share of the head's in-image attention at the peak layer, averaged over heads. It therefore differs slightly from the share in \autoref{tab:sink-families-full}, which is read from the head-averaged map. The \emph{readout} columns divide the contribution norm by the full output norm, for the visual sinks and for the text sink $t^\star$. A readout above 1 means the other contributions partly cancel the sinks'. The \emph{cos} column is the cosine between the output with and without the sink contribution. The three sink-free models are below the rule, with their tokens above $100\times$ uniform in place of the sinks.}
\label{tab:sink-value}
\footnotesize
\begin{tabular}{lcccc}
\toprule
Model & attention & readout: sinks & readout: $t^\star$ & cos \\
\midrule
Qwen3-VL-8B & .70 & .11 & .66 & .994 \\
Qwen3-VL-4B & .75 & .13 & .52 & .991 \\
InternVL3.5-8B & .29 & .03 & .38 & .999 \\
Pixtral-12B & .77 & .16 & .49 & .983 \\
Gemma-4-12B & .62 & 1.20 & .00 & .67 \\
Gemma-4-E4B & .65 & 1.00 & .00 & .41 \\
Gemma-4-E2B & .80 & 1.07 & .00 & .02 \\
\midrule
Qwen3-VL-2B & .10 & .03 & .55 & .999 \\
InternVL3.5-2B & .16 & .03 & .43 & .999 \\
InternVL3-8B & .02 & .01 & .41 & 1.000 \\
\bottomrule
\end{tabular}
\end{table}

In the full-attention families the sinks take $.29$ to $.77$ of the in-image attention but supply $.03$ to $.16$ of the output norm. Removing their contribution barely rotates the output (cosine at least $.98$). HR-Bench gives the same picture (\S~\ref{app:sink-hr}). On the three sink-free models the tokens above $100\times$ uniform take $.02$ to $.16$ of the in-image attention and $.01$ to $.03$ of the output norm. A FastV-style pruner \citep{chen2024fastv} ranks image tokens by attention at layer 20 and keeps the top half or quarter from layer 21 on. Ranked on the raw map it keeps the sinks, and ranked on the cleaned map at one quarter it drops them. Qwen3-VL-8B's V$^\ast$ accuracy is $.73$ in every setting with perfect answer agreement, so dropping the sinks outright changes nothing either.

Gemma-4's sliding-window layers cannot reach the \tsink, and its \vsinks have values of ordinary size. Their readout is $1.00$ to $1.20$, so they supply the dominant share of the output. Removing them rotates the output (cosine $.02$ to $.67$). The same attention pattern is inert in the full-attention families and load-bearing in Gemma-4.

\subsection{Masking instead of editing}
\label{app:sink-mask}

If the sinks are just the most-attended image tokens, can a reader of the map drop them without touching the model? \autoref{tab:sink-mask} tests this on the raw maps. At each upper-half layer we zero every image token above $100\times$ uniform (or above $10\times$) without knowledge of the target box. We then recompute top-1 localization on 100 V$^\ast$ images. In the Qwen3 backbones the mask recovers less than half of the key edit's gain. The sink-free models lose localization under the mask. There $25\%$ to $86\%$ of the tokens above the threshold lie inside the target box (in-box column), so masking them masks the target. The key edit moves the sink-free models by at most $.05$. The $10\times$ threshold is worse on every model but InternVL3.5-8B, since target tokens sit at $10$ to $100\times$ uniform. On Gemma-4 every column is near zero because most upper-half layers use a sliding window and never localize. Over its full-attention layers alone the raw map localizes on $.08$ to $.17$ of images. The $100\times$ mask changes this by at most $.05$, the $10\times$ mask lowers it to $.04$ to $.08$, and the key edit raises it to $.17$ to $.31$. The re-look does not edit Gemma-4's keys (\S~\ref{sec:selector}).

\begin{table}[h]
\centering
\caption{Share of V$^\ast$ images whose most-attended image token lies in the target box. The columns give the raw map, the map after masking (every token above $100\times$ or above $10\times$ uniform zeroed) and the map after the key edit. \emph{In box} is the share of the $>100\times$ tokens inside the target. Numbers average the upper-half layers on 100 images, so the key-edit column is not the value of \autoref{tab:sink-removal}. The Gemma-4 rows average all upper-half layers, sliding-window layers included.}
\label{tab:sink-mask}
\footnotesize\setlength{\tabcolsep}{6pt}
\begin{tabular}{lccccc}
\toprule
Model & raw & masked $>100\times$ & masked $>10\times$ & key edit & in box \\
\midrule
Qwen3-VL-8B & .00 & .20 & .10 & .45 & 5\% \\
Qwen3-VL-4B & .00 & .17 & .10 & .39 & 3\% \\
InternVL3.5-8B & .00 & .18 & .28 & .66 & 2\% \\
Pixtral-12B & .00 & .09 & .03 & .14 & 0\% \\
\midrule
Qwen3-VL-2B & .22 & .18 & .05 & .17 & 25\% \\
InternVL3.5-2B & .57 & .47 & .05 & .56 & 86\% \\
InternVL3-8B & .62 & .50 & .05 & .61 & 85\% \\
\midrule
Gemma-4-12B & .03 & .04 & .10 & .03 & 1\% \\
Gemma-4-E4B & .03 & .03 & .02 & .04 & 0\% \\
Gemma-4-E2B & .02 & .02 & .02 & .07 & 0\% \\
\bottomrule
\end{tabular}
\end{table}

\subsection{Redistribution instead of removal}
\label{app:sink-redistribution}

Does forcing the sinks' attention onto the image help the answer? The redistribution of \citet{kang2025see} keeps the sinks and moves their attention. At every language-model layer during prefill the attention on gated tokens is spread over the remaining image tokens in proportion to their own attention. We run two variants, one in all heads and one only in heads that put more than half their attention on the image. On Qwen3-VL-8B this raises absolute target attention by up to $2.6\times$ and restores localization at the layers that hold the sinks. InternVL3.5-8B's target attention doubles or triples. Accuracy stays flat on both models for both variants. Agreement ($.89$ to $.97$) and KL ($.01$ to $.09$) show that redistribution disturbs the output more than the key edit.

\subsection{The edit along a chain of thought}
\label{app:sink-invariance}

\begin{figure}[h]
\centering
\includegraphics[width=\linewidth]{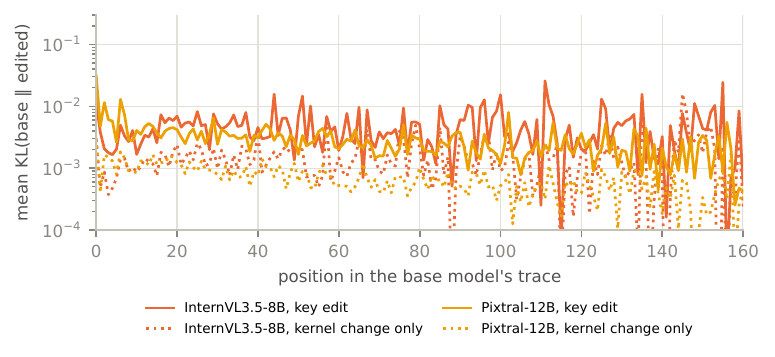}
\caption{Per-position KL along the base model's chain-of-thought trace with the key edit applied at every step (solid) against a kernel-only control (dotted). InternVL3.5-8B and Pixtral-12B on 60 V$^\ast$ questions, greedy chains of up to 192 tokens.}
\label{fig:sink-kl}
\end{figure}

Does the first-token KL of \autoref{tab:sink-removal} stay small along a full trace? We generate greedy chains of thought of up to 192 tokens from InternVL3.5-8B and Pixtral-12B on 60 V$^\ast$ questions with the edit applied at every step. Each trace is scored against the base model's trace. The control is the unedited model rerun with SDPA in place of the eager attention kernel. Per position the teacher-forced KL stays at a few $10^{-3}$ nats along the whole trace (\autoref{fig:sink-kl}). This is an order of magnitude above the kernel control and shows no upward trend.

\subsection{The 2B models}
\label{app:sink-ablations}

The key edit costs Qwen3-VL-2B 7 points on V$^\ast$ ($.68$ to $.61$, \autoref{tab:sink-removal}) and 13 to 14 points on HR-Bench (\autoref{tab:sink-removal-hr}). InternVL3.5-2B loses 3 points on V$^\ast$. Why does the edit hurt them? The sparsity cap and median target of \S~\ref{sec:keyclean} guard against layers where the whole image-key population carries a small component along $\usink$. Shifting that component can turn repulsion into attraction and create a new sink. Qwen3-VL-2B and InternVL3.5-2B sit at the edge of that guard. In InternVL3.5-2B the gate admits weakly aligned non-sink tokens. Qwen3-VL-2B is hurt more. At layer 2 its gate admits tokens with a small $\usink$ component, and moving them to the median creates a new concentration at nearly $1{,}000\times$ uniform. Restricting the edit on Qwen3-VL-2B to the upper three quarters of depth restores accuracy to $.65$. A second unedited run scores $.66$, so this is within run-to-run variation. Agreement is $.98$ and KL is $.008$. For small models the edit should be depth-restricted.

In the re-look the key edit is applied only in the prefill that selects the crop, and the generation passes run on the unedited model. The accuracy loss above therefore does not enter \autoref{tab:main}. On Qwen3-VL-2B the edit can only move the window.

\subsection{The same measurements on HR-Bench}
\label{app:sink-hr}

Do the V$^\ast$ measurements hold on HR-Bench? We repeat the measurements of \autoref{sec:sink} on the first 50 distinct images of HR-4K and of HR-8K with the same budgets and prompts. The 800 questions of each set share 151 images. HR-Bench has no target boxes, so the box columns are absent and a sink is defined by the $100\times$ threshold alone. \autoref{tab:sink-families-hr} gives the sink measurements and \autoref{tab:sink-removal-hr} the key edit. Detector AUC and value readout match V$^\ast$ to within the rounding of \autoref{tab:sink-detectors} and \autoref{tab:sink-value}. We do not reproduce those two tables for HR-Bench. On the four sink-bearing full-attention models the key edit acts as it does on V$^\ast$. Gemma-4-12B and Qwen3-VL-2B are the exceptions, and the removal paragraph below says how they differ.

\begin{table}[h]
\centering
\caption{Text and visual sinks on the first 50 distinct images of HR-4K and of HR-8K. Columns as in \autoref{tab:sink-families-full} without the LM, position, layer and target columns. The visual-sink columns are read at the peak layer. $n$ is the number of image tokens per prompt.}
\label{tab:sink-families-hr}
\footnotesize\setlength{\tabcolsep}{3.5pt}
\resizebox{\linewidth}{!}{%
\begin{tabular}{l ccccc ccccc}
\toprule
& \multicolumn{5}{c}{HR-4K} & \multicolumn{5}{c}{HR-8K} \\
\cmidrule(lr){2-6}\cmidrule(lr){7-11}
Model & $n$ & $t^\star$ share & $\times$unif. & per image & share & $n$ & $t^\star$ share & $\times$unif. & per image & share \\
\midrule
Qwen3-VL-8B & 12194 & .72 & 4626 & 6.1 & .69 & 12416 & .71 & 4659 & 6.1 & .71 \\
Qwen3-VL-4B & 12194 & .67 & 3468 & 7.6 & .76 & 12416 & .67 & 3675 & 8.1 & .78 \\
Qwen3-VL-2B & 12194 & .74 & 1096 & 5.5 & .17 & 12416 & .75 & 1366 & 6.4 & .20 \\
InternVL3.5-8B & 2555 & .64 & 652 & 2.5 & .41 & 1987 & .65 & 314 & 1.8 & .28 \\
InternVL3.5-2B & 2555 & .76 & 143 & 1.6 & .10 & 1987 & .77 & 110 & 0.9 & .07 \\
Gemma-4-12B & 1088 & .10 & 192 & 4.4 & .62 & 1090 & .10 & 192 & 4.5 & .62 \\
Gemma-4-E4B & 1088 & .21 & 417 & 2.7 & .66 & 1090 & .21 & 397 & 2.7 & .67 \\
Gemma-4-E2B & 1088 & .22 & 525 & 2.7 & .83 & 1090 & .22 & 481 & 2.8 & .82 \\
InternVL3-8B & 2555 & .60 & 126 & 1.1 & .07 & 1987 & .61 & 89 & 0.6 & .04 \\
Pixtral-12B & 3919 & .48 & 3063 & 1.7 & .79 & 2742 & .49 & 2093 & 1.0 & .78 \\
\bottomrule
\end{tabular}}
\end{table}

The sink-bearing split of \autoref{sec:sink} is made on V$^\ast$. On HR-Bench Qwen3-VL-2B's peak share rises to $.17$ and $.20$ and crosses the $15\%$ cut. This goes with the larger accuracy loss of the all-layer edit reported below.

\paragraph{Detectors.}
At the peak layer the centered key cosine keeps near-perfect AUC on every sink-bearing model on both sets. The family-level pattern of precision and of hidden-norm AUC matches \autoref{tab:sink-detectors}.

\paragraph{Value side.}
The sinks' share of the output matches the V$^\ast$ values of \autoref{tab:sink-value} to within $.01$ on every model and dataset. This includes the large Gemma-4 contribution.

\begin{table}[h]
\centering
\caption{The key edit on the first 50 distinct images of HR-4K and of HR-8K. Columns as in \autoref{tab:sink-removal} without the box and caption columns. \emph{Gate off} means the gate admits under one token per image, so the edit is an identity.}
\label{tab:sink-removal-hr}
\footnotesize\setlength{\tabcolsep}{2.5pt}
\resizebox{\linewidth}{!}{%
\begin{tabular}{l ccccc ccccc}
\toprule
& \multicolumn{5}{c}{HR-4K} & \multicolumn{5}{c}{HR-8K} \\
\cmidrule(lr){2-6}\cmidrule(lr){7-11}
Model & $\max_j r_j$ & text sink & accuracy & agreement & KL & $\max_j r_j$ & text sink & accuracy & agreement & KL \\
\midrule
Qwen3-VL-8B & $.25\to.02$ & $.54\to.64$ & $.72\to.72$ & .99 & .001 & $.24\to.02$ & $.54\to.65$ & $.68\to.67$ & .99 & .001 \\
Qwen3-VL-4B & $.21\to.03$ & $.51\to.65$ & $.74\to.74$ & .97 & .000 & $.22\to.03$ & $.51\to.65$ & $.69\to.69$ & .99 & .000 \\
Qwen3-VL-2B & $.02\to.02$ & $.60\to.59$ & $.71\to.57$ & .74 & .017 & $.02\to.02$ & $.61\to.59$ & $.72\to.59$ & .78 & .024 \\
InternVL3.5-8B & $.31\to.04$ & $.42\to.63$ & $.73\to.72$ & .98 & .000 & $.33\to.04$ & $.44\to.64$ & $.69\to.70$ & .99 & .000 \\
InternVL3.5-2B & $.04\to.04$ & $.62\to.61$ & $.76\to.76$ & .87 & .002 & $.04\to.04$ & $.63\to.63$ & $.62\to.63$ & .90 & .008 \\
InternVL3-8B & gate off & $.46\to.46$ & $.78\to.77$ & .99 & .000 & gate off & $.47\to.47$ & $.73\to.74$ & .98 & .001 \\
Pixtral-12B & $.56\to.02$ & $.38\to.49$ & $.61\to.60$ & .95 & .003 & $.59\to.02$ & $.39\to.50$ & $.53\to.52$ & .96 & .002 \\
\midrule
Gemma-4-12B & $.15\to.12$ & -- & $.67\to.68$ & .89 & .002 & $.15\to.14$ & -- & $.54\to.57$ & .77 & .085 \\
Gemma-4-E4B & $.30\to.27$ & -- & $.66\to.62$ & .94 & .001 & $.30\to.27$ & -- & $.58\to.56$ & .95 & .001 \\
Gemma-4-E2B & $.40\to.38$ & -- & $.60\to.59$ & .93 & .000 & $.38\to.36$ & -- & $.54\to.51$ & .93 & .001 \\
\bottomrule
\end{tabular}}
\end{table}

\paragraph{Removal.}
On Qwen3-VL-8B, Qwen3-VL-4B, InternVL3.5-8B and Pixtral-12B the edit clears the map and leaves the answers alone (\autoref{tab:sink-removal-hr}). The freed mass goes to the \tsink, whose share rises by $.10$ to $.21$ on those four models. On InternVL3-8B the gate admits under one token per image. The map does not move on Gemma-4. Its output is close to unchanged on HR-4K (KL at most $.002$, though Gemma-4-E4B loses four points of accuracy) and shifts on HR-8K for Gemma-4-12B (KL $.085$, agreement $.77$). That value is far below the V$^\ast$ KL of $.531$ for the same model. We have not traced this gap. The re-look does not edit Gemma-4's keys (\S~\ref{sec:selector}), so it does not enter \autoref{tab:main}. On Qwen3-VL-2B the degradation grows with input length. The longer sequences admit many more tokens through the gate, and the all-layer edit costs over ten points while the map does not change. The depth-restricted edit of \S~\ref{app:sink-ablations} is the form to use on this model.

\section{Selector Layers}
\label{app:selector-layers}

Which layers carry a question-dependent map? For Qwen3-VL-8B we sweep all 36 layers on 150 images. The images come 30 each from Geo3K, RealWorldQA, V$^\ast$, VMCBench and MMVP. At each layer we compute the question-to-image attention map under the real question and again under an unrelated prompt. The statistic is the overlap of the eight most-attended image tokens between the two maps. A low overlap means the map depends on the question. It is lowest at layers 18 to 21 (we do not plot the sweep). We take the two ends of this band, layers 18 and 21. The choice predates the main experiments and is not tuned per benchmark. Thirty of the 150 sweep images are V$^\ast$ images, so the layer choice has seen part of the V$^\ast$ set. The sweep statistic uses neither the target boxes nor the answers.

The other full-attention models use the same relative depths of about $.5$ and $.58$ and are not searched separately. \autoref{tab:selector-layers} lists the resulting layers for every model.

\begin{table}[h]
\centering
\caption{Selector layers per model. The selector averages the head-averaged question-to-image attention of these two layers (\S~\ref{sec:selector}). The number after the slash is the count of language-model layers. The other full-attention models read the cleaned map at the same relative depths as Qwen3-VL-8B. The Gemma-4 rows list full-attention layers, which the selector reads on the raw map.}
\label{tab:selector-layers}
\footnotesize\setlength{\tabcolsep}{8pt}
\begin{tabular}{ll}
\toprule
Model & layers / total \\
\midrule
Qwen3-VL-8B & 18, 21 / 36 \\
Qwen3-VL-4B & 18, 21 / 36 \\
InternVL3.5-8B & 18, 21 / 36 \\
Qwen3-VL-2B & 14, 16 / 28 \\
InternVL3.5-2B & 14, 16 / 28 \\
InternVL3-8B & 14, 16 / 28 \\
Pixtral-12B & 20, 23 / 40 \\
\midrule
\multicolumn{2}{l}{\emph{Gemma-4: full-attention layers, raw map}} \\
Gemma-4-E2B & 19, 24 \\
Gemma-4-E4B & 17, 23 \\
Gemma-4-12B & 23, 29 \\
\bottomrule
\end{tabular}
\end{table}

On Gemma-4 the selector reads only full-attention layers (\S~\ref{sec:selector}). These lie five or six layers apart, so relative depth does not carry over. We take the adjacent pair of full-attention layers whose crop hits the target most often on 60 V$^\ast$ images. These 60 images are part of the V$^\ast$ evaluation set, so Gemma-4's V$^\ast$ column in \autoref{tab:main} is not fully held out. Its HR-Bench and MME-RealWorld-Lite columns are.

\begin{figure}[t]
\centering
\includegraphics[width=\linewidth]{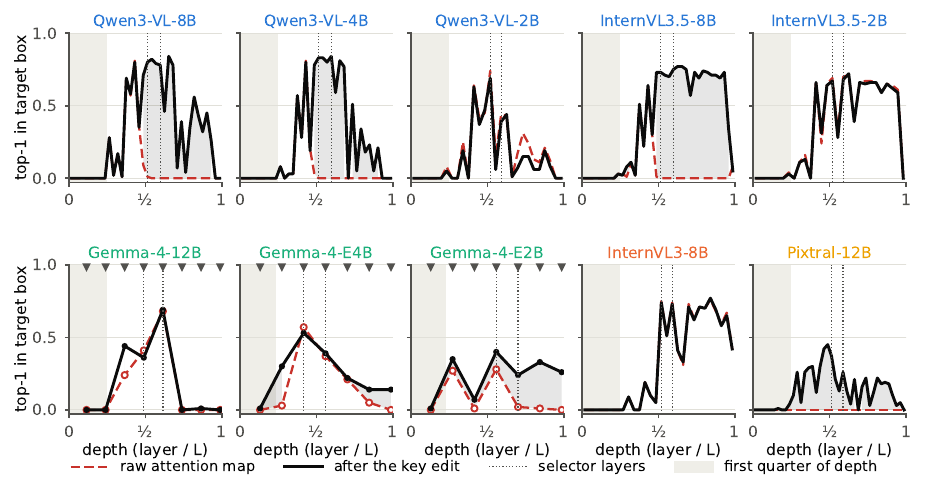}
\caption{Per-layer top-1 localization on V$^\ast$ for the raw map (dashed) and the cleaned map (solid). Here top-1 localization is the fraction of images whose most-attended image token lies in the target box. The shaded area between the curves is the gain from the key edit. Dotted lines mark the selector layers. The gray band is the first quarter of depth, which the peak-layer rule skips. Gemma-4 shows full-attention layers only (triangles).}
\label{fig:sink-removal-grounding}
\end{figure}

Does the cleaned map localize at the chosen layers? \autoref{fig:sink-removal-grounding} plots top-1 localization per layer on V$^\ast$ for the raw and the cleaned map, with the selector layers marked. Every full-attention model but Pixtral-12B has a selector layer at which the cleaned map puts its most-attended token in the target box on $.68$ to $.84$ of V$^\ast$ images. This is within $.05$ of the model's best layer past the first quarter of depth. Pixtral-12B reaches $.37$ at layer 20 and $.45$ at its best layer, layer 19. These are per-layer values read from \autoref{fig:sink-removal-grounding}. \autoref{tab:sink-removal} averages over the probed layers, which is why its box column is lower.

After the key edit the most-attended token still lands on the outermost row or column of the grid several times more often than chance in the late layers of every model. The effect is strongest in Pixtral-12B, where the tokens next to \texttt{[IMG\_BREAK]} keep an alignment just under the gate threshold at every layer past mid depth. At the Qwen3-VL selector layers the border rate is at chance. No run in this paper corrects for this effect.

\section{Experimental Details}
\label{app:exp-details}

\paragraph{Models and prompts.}
Every model receives the image followed by the question and its lettered options, verbatim from the benchmark. One answer-format sentence closes the prompt and depends on how the model reasons. The native thinking models are asked only for the letter. They reason in their own thinking channel. Qwen3-VL-Thinking uses its default chat template with no system prompt. For InternVL3.5 the R1 system prompt of its model card switches on the thinking channel. Gemma-4 runs through its template with thinking enabled. InternVL3-8B and Pixtral-12B have no thinking channel and are asked to think step by step in the answer. MME-RealWorld-Lite has five options (A to E) and the same wording. \autoref{fig:prompts} gives the exact strings.

\begin{figure*}[t]
    \noindent\fbox{\parbox{.98\textwidth}{\small\ttfamily\raggedright\setlength{\parskip}{4pt}
    \textbf{\rmfamily Thinking models:} \{image\} \{question\}\textbackslash nPut only the letter of the correct option inside \textbackslash boxed\{\}.

    \par\textbf{\rmfamily Chain-of-thought prompted models:} \{image\} \{question\}\textbackslash nThink step by step, then put only the letter of the correct option inside \textbackslash boxed\{\}.

    \par\textbf{\rmfamily Transition after the spliced crop:} \{crop\} Looking at this region again,
    }}
    \caption{Prompts used in every run. The native thinking models receive the first line. Qwen3-VL-Thinking (8B, 4B and 2B) uses its default chat template with no system prompt. For InternVL3.5 (8B and 2B) the R1 system prompt of its model card switches on the thinking channel. Gemma-4 (12B, E4B and E2B) runs its default chat template with \texttt{enable\_thinking=True}. InternVL3-8B and Pixtral-12B have no thinking channel and receive the second line. The third line follows the spliced crop.}
    \label{fig:prompts}
\end{figure*}

\paragraph{Image budgets.}
Qwen3-VL-8B, Qwen3-VL-4B and Qwen3-VL-2B see at most 16{,}384 patches of $28^2$ pixels (12.8 Mpix). InternVL3.5-8B, InternVL3.5-2B and InternVL3-8B keep their default dynamic tiling of up to 12 tiles of $448^2$ pixels. Pixtral-12B uses its processor default of a 1{,}024-pixel longest edge. Gemma-4 runs at its maximum of 1{,}120 soft tokens. HR-4K and HR-8K images have medians of 16 and 39 Mpix, against 3.4 Mpix on V$^\ast$.

\paragraph{Sampling and trigger.}
Sampling uses temperature $0.6$ with top-$p$ $0.95$, top-$k$ $30$ and min-$p$ $0.001$. Generation is capped at 4{,}096 new tokens. The seed is the question index. The re-look is triggered at entropy $H_0=1$ nat after at least $c=32$ tokens of thought (\S~\ref{sec:trigger}). The selector reads two mid-depth layers per model (\autoref{app:selector-layers}) with the key edit at $\tau_c=0.5$ and $\rho=0.02$ (\S~\ref{sec:keyclean}). The edit is applied only in the prefill that selects the crop. On Gemma-4 the selector reads two full-attention layers without the edit. The window is a third of the grid on each side, padded by 20 pixels.

\paragraph{Trigger coverage.}\label{app:trigger-coverage}
\autoref{tab:trigger-coverage} gives the per-model trigger rates. For six of the ten models the trigger fires on more than 90\% of questions on every benchmark. On those models the method is close to an unconditional re-look. We do not report an always-trigger or random-step control.

\begin{table}[h]
  \caption{Fraction of questions (\%) on which the re-look is triggered, per model and benchmark. V$^\ast$ has 191 questions, HR-4K and HR-8K 800 each, MME-RealWorld-Lite 1{,}919. Same runs as \autoref{tab:main}.}
  \label{tab:trigger-coverage}
  \begin{center}
  \footnotesize
  \begin{tabular}{l cccc}
    \toprule
    \textbf{Model} & V$^\ast$ & HR-4K & HR-8K & MME-RW-Lite \\
    \midrule
    \multicolumn{5}{l}{\textit{Qwen3-VL-Thinking}} \\
    \quad 8B & 59.7 & 60.1 & 73.1 & 90.9 \\
    \quad 4B & 65.4 & 65.4 & 74.9 & 94.2 \\
    \quad 2B & 72.3 & 71.8 & 78.2 & 92.5 \\
    \addlinespace[2pt]
    \multicolumn{5}{l}{\textit{InternVL3.5 (thinking)}} \\
    \quad 8B & 79.6 & 79.9 & 83.9 & 91.0 \\
    \quad 2B & 100.0 & 100.0 & 100.0 & 100.0 \\
    \addlinespace[2pt]
    \multicolumn{5}{l}{\textit{Gemma-4 (thinking)}} \\
    \quad 12B & 99.5 & 98.2 & 99.1 & 99.6 \\
    \quad E4B & 99.5 & 99.1 & 99.8 & 99.3 \\
    \quad E2B & 97.9 & 98.4 & 98.6 & 99.0 \\
    \addlinespace[2pt]
    \multicolumn{5}{l}{\textit{Chain-of-thought prompted}} \\
    \quad InternVL3-8B & 96.3 & 97.9 & 97.5 & 98.3 \\
    \quad Pixtral-12B & 95.3 & 91.4 & 93.0 & 96.5 \\
    \bottomrule
  \end{tabular}
  \end{center}
\end{table}

\paragraph{Paired comparisons.}\label{app:paired}
\autoref{tab:paired} gives the per-question counts behind the gains of \autoref{tab:main}. Each cell counts the discordant questions between the re-look on the cleaned map and a reference run. Against the base run the re-look wins 1{,}997 discordant questions and loses 1{,}181 over the 30 (model, benchmark) pairs on V$^\ast$ and HR-Bench, a ratio of 1.7. Against the raw map the cleaned map is significantly better in seven of the 28 pairs that have a raw run and never significantly worse. Five of the seven are on V$^\ast$ and HR-Bench. The last two columns show where the crops land on V$^\ast$. A crop counts as a hit when it contains the center of a target box. On Qwen3-VL-8B the cleaned map hits on .90 of triggered questions and the raw map on .22. On the three sink-free models the raw map hits as often as the cleaned map or slightly more often (.82 against .76 on Qwen3-VL-2B).

\begin{table}[t]
  \caption{Paired comparisons behind \autoref{tab:main}. For every model and benchmark, wins/losses counts the questions the re-look with the cleaned map answers correctly and the reference answers wrongly, against the reverse. The reference is the base run (\emph{base}) or the same re-look with the crop selected on the raw map (\emph{raw}). $^{\dagger}$ marks $p<.05$ under the exact McNemar test. The last two columns give the V$^\ast$ crop-hit rate on triggered questions for the cleaned and for the raw map. A crop hits when it contains the center of a target box. Gemma-4's selector reads the raw map, so its one hit rate is listed under \emph{cleaned} and it has no raw columns. V$^\ast$ has 191 questions, HR-4K and HR-8K 800 each, MME-RealWorld-Lite 1{,}919. Same runs as \autoref{tab:main}.}
  \label{tab:paired}
  \begin{center}
  \footnotesize\setlength{\tabcolsep}{3.5pt}
  \resizebox{\textwidth}{!}{%
  \begin{tabular}{l cc cc cc cc cc}
    \toprule
    & \multicolumn{2}{c}{\textbf{V$^\ast$}} & \multicolumn{2}{c}{\textbf{HR-4K}} & \multicolumn{2}{c}{\textbf{HR-8K}} & \multicolumn{2}{c}{\textbf{MME-RW-Lite}} & \multicolumn{2}{c}{\textbf{V$^\ast$ crop hit}} \\
    \cmidrule(lr){2-3} \cmidrule(lr){4-5} \cmidrule(lr){6-7} \cmidrule(lr){8-9} \cmidrule(lr){10-11}
    \textbf{Model} & vs.\ base & vs.\ raw & vs.\ base & vs.\ raw & vs.\ base & vs.\ raw & vs.\ base & vs.\ raw & cleaned & raw \\
    \midrule
    \multicolumn{11}{l}{\textit{Qwen3-VL-Thinking}} \\
    \quad 8B & 9/2 & 7/4 & 51/18$^{\dagger}$ & 40/13$^{\dagger}$ & 65/30$^{\dagger}$ & 41/22$^{\dagger}$ & 114/75$^{\dagger}$ & 83/64 & .90 & .22 \\
    \quad 4B & 11/6 & 13/4$^{\dagger}$ & 30/23 & 32/21 & 62/21$^{\dagger}$ & 59/30$^{\dagger}$ & 144/79$^{\dagger}$ & 118/68$^{\dagger}$ & .89 & .08 \\
    \quad 2B & 14/11 & 4/7 & 58/34$^{\dagger}$ & 12/12 & 71/46$^{\dagger}$ & 16/24 & 159/120$^{\dagger}$ & 38/40 & .76 & .82 \\
    \addlinespace[2pt]
    \multicolumn{11}{l}{\textit{InternVL3.5 (thinking)}} \\
    \quad 8B & 23/6$^{\dagger}$ & 5/2 & 65/16$^{\dagger}$ & 17/18 & 86/30$^{\dagger}$ & 10/16 & 172/113$^{\dagger}$ & 95/77 & .88 & .80 \\
    \quad 2B & 34/14$^{\dagger}$ & 4/3 & 108/74$^{\dagger}$ & 8/11 & 142/66$^{\dagger}$ & 12/9 & 189/181 & 57/51 & .81 & .81 \\
    \addlinespace[2pt]
    \multicolumn{11}{l}{\textit{Gemma-4 (thinking)}} \\
    \quad 12B & 22/9$^{\dagger}$ & -- & 88/51$^{\dagger}$ & -- & 122/43$^{\dagger}$ & -- & 200/147$^{\dagger}$ & -- & .84 & -- \\
    \quad E4B & 38/17$^{\dagger}$ & -- & 96/81 & -- & 108/68$^{\dagger}$ & -- & 216/146$^{\dagger}$ & -- & .75 & -- \\
    \quad E2B & 33/22 & -- & 83/73 & -- & 127/74$^{\dagger}$ & -- & 182/191 & -- & .50 & -- \\
    \addlinespace[2pt]
    \multicolumn{11}{l}{\textit{Chain-of-thought prompted}} \\
    \quad InternVL3-8B & 21/10 & 0/0 & 85/82 & 1/0 & 121/78$^{\dagger}$ & 2/2 & 226/208 & 9/9 & .91 & .91 \\
    \quad Pixtral-12B & 30/16 & 13/7 & 83/74 & 60/34$^{\dagger}$ & 111/86 & 58/38 & 253/182$^{\dagger}$ & 152/107$^{\dagger}$ & .38 & .13 \\
    \bottomrule
  \end{tabular}}
  \end{center}
\end{table}

\end{document}